\documentclass{article}
\PassOptionsToPackage{numbers, compress}{natbib}
\usepackage[main, final]{neurips_2026}
\usepackage[T1]{fontenc}    
\usepackage[utf8]{inputenc} 
\usepackage{amsfonts}       
\usepackage{amsmath}
\usepackage{amssymb}
\usepackage{bm}
\usepackage{booktabs}
\usepackage{colortbl}
\usepackage{enumitem}
\usepackage{float}
\usepackage{graphicx}
\usepackage{hyperref}       
\usepackage{mathtools}
\usepackage{microtype}
\usepackage{multirow}
\usepackage{placeins}
\usepackage{needspace}
\usepackage{nicefrac}       
\usepackage{subcaption}
\usepackage{url}            
\usepackage{wrapfig}
\usepackage{xcolor}         
\definecolor{best}{RGB}{180, 220, 200}
\definecolor{second}{RGB}{225, 250, 240}
\makeatletter
\newcommand{\oursrowshade}{\rlap{\color{gray!10}\vrule width\linewidth height\ht\@arstrutbox depth\dp\@arstrutbox}}
\makeatother

\usepackage[accsupp]{axessibility}
\usepackage{hyperref}
\hypersetup{hidelinks}
\usepackage[capitalize,noabbrev,nameinlink]{cleveref}
\crefname{section}{Section}{Sections}
\Crefname{section}{Section}{Sections}
\crefname{subsection}{Section}{Sections}
\Crefname{subsection}{Section}{Sections}
\crefname{appendix}{Appendix}{Appendices}
\Crefname{appendix}{Appendix}{Appendices}
\usepackage{orcidlink}

\usepackage{xspace}

\makeatletter
\DeclareRobustCommand\onedot{\futurelet\@let@token\@onedot}
\def\@onedot{\ifx\@let@token.\else.\null\fi\xspace}
\def\eg{\emph{e.g}\onedot}

\def\vs{\emph{vs}\onedot}

\makeatother

\newcommand{\nickname}{\textsc{Track4D}\xspace}

\newcommand{\ncmv}{\textsc{NCMV}}
\newcommand{\rtv}{\textsc{RTV}}

\PassOptionsToPackage{pagebackref}{hyperref}

\newcommand{\heading}[1]{\vspace{2pt}\noindent\textbf{#1.}}

\newcommand{\pzero}[2]{P_{#1}\!\left(t_{#2}\right)} 

\DeclareRobustCommand{\corange}[1]{\textcolor{orange}{#1}}

\definecolor{tealborder}{RGB}{5,170,159}
\DeclareRobustCommand{\cteal}[1]{\textcolor{tealborder}{#1}}

\makeatletter
\renewcommand{\paragraph}{%
  \@startsection{paragraph}{4}%
  {\z@}{-0.5em}{-0.5em}%
  {\normalfont\normalsize\bfseries}%
}
\makeatother

\title{\nickname: Representing Dense 3D Tracking for Video Diffusion Models}

\author{%
\begin{tabular}{@{}c@{\hspace{2.5em}}c@{\hspace{2.5em}}c@{}}
Yushi Lan$^{1}$ & Zeren Jiang$^{1}$ & Kelvin Li$^{1}$\\[0.4em]
Xingang Pan$^{2}$ & Chuanxia Zheng$^{2}$ & Andrea Vedaldi$^{1}$
\end{tabular}\\
\rule{0pt}{1.8em}
$^{1}$Visual Geometry Group, University of Oxford \quad
$^{2}$Nanyang Technological University\\[0.3em]
{\footnotesize \texttt{\{yushi,zeren,kelvin,vedaldi\}@robots.ox.ac.uk \quad \{xingang.pan,chuanxia.zheng\}@ntu.edu.sg}}\\[0.3em]
\href{https://nirvanalan.github.io/projects/track4d}{\textcolor{magenta}{\texttt{https://nirvanalan.github.io/projects/track4d}}}
}

\begin{document}
\maketitle
\raggedbottom

\setlength{\textfloatsep}{7pt plus 2pt minus 2pt}
\makeatletter\setlength{\@neuripsabovecaptionskip}{4pt}\setlength{\abovecaptionskip}{4pt}\makeatother
\suppressfloats[t]
\begin{figure}[h!]
\vspace{-18pt}
\centering
\includegraphics[width=\textwidth]{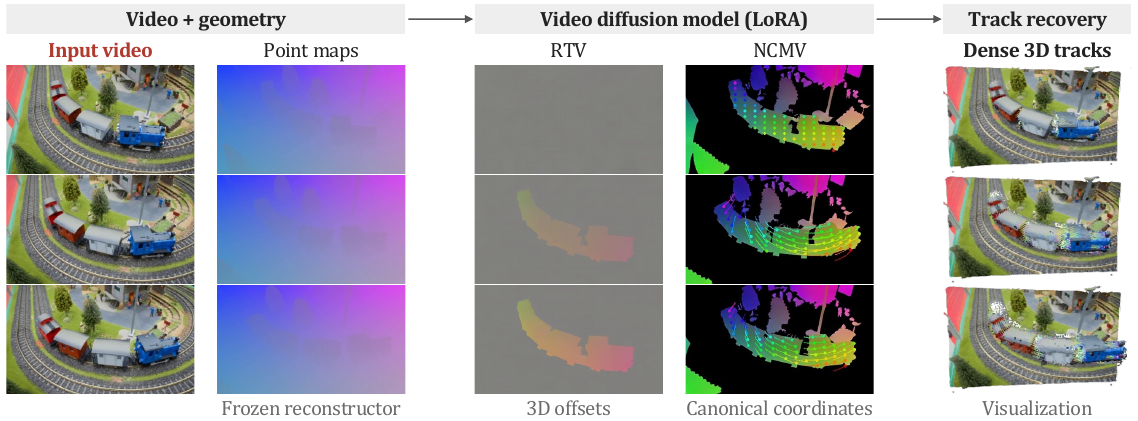}
\caption{\textbf{\nickname~formulates dense 3D tracking as conditional video generation.}
From a monocular video and point maps of a frozen 3D reconstructor, a LoRA-adapted video diffusion model generates a Residual 3D Tracking Video (metric 3D offsets) and a Normalized Coordinate Map Video (canonical coordinates), from which dense 3D tracks are recovered without per-scene optimization.}%
\label{fig:teaser}
\end{figure}
\begin{abstract}
We study how to repurpose video diffusion models (VDMs), pretrained on millions of videos, for dense 3D point tracking, where training data is scarce.
Doing so is non-trivial, as it requires representing 3D tracks as video-like signals that may lie outside the natural-video distribution on which the VDM was trained.
We thus systematically assess the effectiveness of two such video representations.
The first, Residual 3D Tracking Video (\rtv), encodes per-pixel 3D offsets relative to a query frame and allows recovery of 3D trajectories directly from the generated video.
The second, Normalized Coordinate Map Video (\ncmv), labels each pixel with a canonical coordinate invariant through time.
This is closer to a natural video and easier for the VDM to learn, but requires additional post-processing with an off-the-shelf 2D tracker to recover 3D tracks.
\rtv{} tracks more accurately, while \ncmv{} shows that a VDM can generate dense correspondence as a video.
We instantiate both representations with LoRA adaptation of a frozen Wan backbone, conditioned on point maps from an off-the-shelf 3D reconstructor as a geometric scaffold.
Trained with a single dynamic 3D tracking dataset, \nickname{} achieves competitive zero-shot tracking on four real and synthetic benchmarks.
A regression readout that bypasses the video VAE and uses additional training data brings further improvement, demonstrating the potential of this direction for dense 3D tracking.
We hope our analysis will inspire future work that leverages VDM priors for dense 3D tracking and 4D reconstruction.
More details can be found on our \href{https://nirvanalan.github.io/projects/track4d}{\textcolor{magenta}{project page}}.
\end{abstract}

\section{Introduction}%
\label{sec:intro}

We consider the problem of \emph{dense monocular 4D reconstruction}, i.e., recovering the 3D shape and motion of every visible surface point in a video.
4D reconstruction is propaedeutic to understanding the physical world and supports applications such as video editing~\cite{lee2025editbytrack}, spatial reasoning~\cite{xiao2025video4spatial} and robotics~\cite{lee2025tracegen,huang2026pointworldscaling3dworld}.
Traditional 4D reconstruction relied on per-scene optimization~\cite{torresani08nonrigid,schoenberger2016sfm,wang2025shape,rajivc2025multi}, which is costly and brittle as dynamic reconstruction needs strong \emph{statistical priors}~\cite{torresani08nonrigid}.
Feed-forward networks now excel at \emph{static} 3D reconstruction~\cite{wang2024dust3r,wang2025cut3r,wang25vggt,wang2025pi,stream3r2025,lin2026depth,wang26vggt-omega};
for \emph{dynamics}, sparse 3D point trackers~\cite{xiao2024spatialtracker,xiao2025spatialtracker,tapip3d} scale poorly to dense outputs, and dense 4D methods~\cite{st4rtrack2025,jin2025stereo4d,sucar25dynamic,sucar2025vdpm,karhade2025any4d,zhang2025d4rt} are bottlenecked by the scarcity of video with \emph{dense 3D tracking annotations}.

Here we propose to address the scarcity of training data by leveraging Video Diffusion Models (VDMs)~\cite{videoworldsimulators2024,wan2025} pretrained on millions of unlabelled videos.
VDMs have already been shown to transfer well to depth prediction~\cite{ke2024repurposing}, dynamic 3D reconstruction~\cite{geo4d}, and short-term correspondences~\cite{zhu2026motioncrafter}.
How to extend VDMs to 3D tracking is, however, non-trivial, as one must represent the 3D tracks as a video-like signal, representing at every pixel the location of the corresponding 3D point in world space.
While this signal may be in the \emph{form} of a video, i.e., a spatio-temporal map, it \emph{looks nothing like a natural video}.
How to adapt the latent space of the VDM to represent such a signal is the key engineering problem.
In this paper, we explore this design choice systematically.

We note that there are two fairly different ways to represent 3D tracks as a video (\cref{fig:teaser}).
The first is a \emph{Residual 3D Tracking Video} (\rtv), where each pixel in a given query frame is mapped over time to different 3D displacement vectors (offsets).
By looking at how the value associated to a pixel varies over time, one recovers directly the 3D trajectory of the corresponding physical point, tracking its motion forward.
However, as shown in~\cref{fig:teaser}, the resulting video is unnatural as motion is encoded as an intensity change rather than a spatial shift.
The other approach is to use instead a \emph{Normalized Coordinate Map Video} (\ncmv), which associates each pixel to the 3D coordinate of the corresponding canonical physical point at a fixed time and spatial reference frame.
These canonical coordinates assign to each physical point a unique `color', consistent over time.
The resulting video is more natural, as motion is encoded as a spatial shift of these colors, and is thus easier for the VDM to generate.
However, recovering the forward 3D trajectories requires further 2D tracking.

We explore both representations, using LoRA to adapt a frozen Wan~\cite{wan2025} VDM backbone for 3D tracking (\cref{fig:two-rep=for-vdm}).
We condition the generator on the output of an off-the-shelf 3D reconstructor (\eg VGGT~\cite{wang25vggt}, STream3R~\cite{stream3r2025}, DAv3~\cite{lin2026depth}), providing a geometric scaffold to the model, and learn to decode 3D tracks by fine-tuning the VDM\@. Since only LoRA adapters are trained, \nickname{} directly benefits from advances in video generation, and even the 14B backbone can be fine-tuned with academic resources.

Encoding tracks as a video requires a target at every pixel, which restricts training to a single dense dynamic 3D tracking dataset (Kubric MOVi-F~\cite{greff2021kubric}). Nevertheless, \nickname achieves competitive zero-shot performance on four tracking benchmarks and, on the held-out dense Kubric split, outperforms St4RTrack~\cite{st4rtrack2025} and TraceAnything~\cite{liu2025traceanythingrepresentingvideo} by a clear margin in dense APD/EPE\@.
As for the best representation, we find that \rtv{} is simpler and results in better forward tracking performance, despite arguably being harder to learn.
Still, \ncmv{} shows that a VDM can generate dense correspondence directly as a video output, whereas prior work only reads sparse 2D tracks from \emph{frozen} VDM features~\cite{tang2023dift,nam2025difftrack}. This suggests that video generators carry enough structure to produce dense correspondence, even if \ncmv{} is not the most accurate way to read out tracks.

To summarise, we make three contributions. First, we formulate dense 3D tracking as conditional video generation in a pretrained VDM, distinct from prior work~\cite{tang2023dift,nam2025difftrack} that reads sparse 2D correspondences from \emph{frozen} VDM features.
This allows us to \emph{sample} trajectories from the VDM's generative prior instead of reading them out deterministically; although VDMs denoise iteratively, few-step denoising retains similar accuracy at a lower cost (\cref{tab:sampling_steps}).
Second, we systematically study how to represent 3D tracks inside a VDM, comparing \rtv{} (explicit 3D offsets) and \ncmv{} (implicit canonical-coordinate correspondence) within a single model: \rtv{} is the recommended output, while \ncmv{} serves as a probe of VDM correspondence priors and as an auxiliary training objective, since the joint checkpoint slightly improves \rtv{} (\cref{appendix:ncmv-analysis}).
Third, we study a VAE-free regression readout that admits sparse supervision and improves further with more data.
Since it learns a residual on the input point map, its accuracy is limited by that geometry, which suggests training geometry and motion jointly~\cite{sucar2025vdpm,luo264rc:}, left to future work (\cref{sec:regression-results}).

\begin{figure}[t]
\centering
\includegraphics[width=\textwidth]{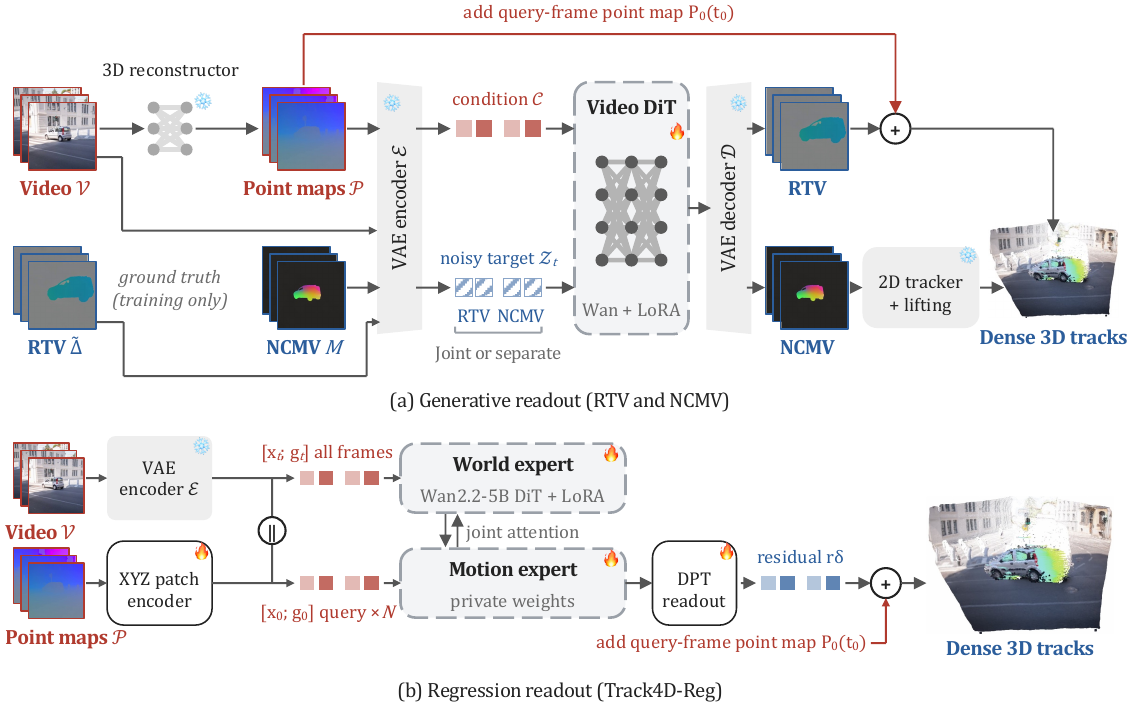}
\caption{\textbf{\nickname~overview.}
\textbf{(a) Generative readout.} A frozen reconstructor supplies point maps $\mathcal{P}$; the video VAE encodes them with video $\mathcal{V}$ into condition $\mathcal{C}$.
The LoRA-adapted video DiT denoises the \rtv{} and \ncmv{} latents, jointly by default, and the VAE decodes the videos. Hatched targets are noised in training and start from Gaussian noise at inference.
Adding inverse-normalized \rtv{} offsets to the query-frame point map gives 3D tracks; \ncmv{} requires 2D tracking and lifting.
\textbf{(b) Regression readout.} Point maps enter through a patch encoder, a private motion expert attends jointly with the video DiT, and a DPT head reads residuals without the output VAE (\cref{sec:regression,fig:supp-mot-arch}).
Flames denote trained modules, snowflakes frozen modules.}%
\label{fig:method_overview}
\end{figure}

\section{Related work}%
\label{sec:related}

\heading{Feed-forward 3D and 4D reconstruction}
Predicting dense 3D geometry in a single forward pass has progressed from pairwise point map regression~\cite{wang2024dust3r} replacing classical pipelines~\cite{schoenberger2016sfm,schoenberger2016mvs,yao2018mvsnet,yao2019recurrent}, to scalable static multi-view models~\cite{wang25vggt,lin2026depth,yang25fast3r:,wang26vggt-omega} and dynamic sequential methods~\cite{wang2025cut3r,stream3r2025,xu2025dgt,liang2025btimer}.
St4RTrack~\cite{st4rtrack2025} and DPM~\cite{sucar25dynamic} pioneered dense 4D reconstruction by augmenting DUSt3R-style architectures with motion prediction; subsequent work~\cite{liu2025traceanythingrepresentingvideo,sucar2025vdpm,karhade2025any4d,zhang2025d4rt,zhang25pomato:,liang25zeromsf,lu26track4world,luo264rc:} further advances this direction.
Because these models build on 3D reconstructors trained on static data, extending them to 4D requires tracking supervision, which remains scarce~\cite{greff2021kubric,zheng2023point,karaev2023dynamicstereo}.
These deterministic regressors optimize geometry and motion directly and can also supply our geometric scaffold (\cref{appendix:recent-methods}).
\nickname{} takes a complementary route by adapting priors already encoded in a pretrained VDM.

\heading{Dense 3D point tracking}
Temporal correspondence has been studied through optical flow~\cite{sun2010secrets,sun2018pwc,teed2020raft} and long-range 2D point tracking~\cite{pips,tapvid,tapir,cotracker,harley2025alltracker}.
Lifting 2D tracks to 3D usually requires a separate depth model: SpatialTracker~\cite{xiao2024spatialtracker,xiao2025spatialtracker} and TAPIP3D~\cite{tapip3d} follow this two-stage strategy but inherit errors from each stage.
DELTA~\cite{ngo2024delta,ngo2025deltav2} trains end-to-end dense 3D trackers with sparse-to-dense upsampling but require curated tracking supervision.
The generative \nickname{} differs in two ways: it produces dense 3D tracks directly as VDM output rather than from a supervised 3D-tracking head, and it combines pretrained motion priors with limited tracking supervision.
The introduced representation \rtv{} is closely related to neural scene-flow representations~\cite{li21nsfp:,wang2025shape}, but is dense, in metric world coordinates, and generated through iterative denoising without per-pair optimization.

\heading{Repurposing video diffusion models for geometric tasks}
Pretrained video generators~\cite{ho22video,blattmann23align,blattmann23stable,wan2025,videoworldsimulators2024,kong2024hunyuanvideo} encode powerful spatio-temporal priors.
On the image side, Marigold~\cite{ke2024repurposing} and follow-ups~\cite{xu2025pixel} fine-tune diffusion models for depth.
For dynamic scenes, Geo4D~\cite{geo4d} predicts per-frame point maps, 4DNex~\cite{chen20254dnex} adopts width-concatenated multi-modal latents, and VIST3A~\cite{go2026texttod} and Gen3R~\cite{huang2026gen3r3dscenegeneration} couple a VGGT backbone with a VDM for joint RGB and 3D generation.
However, these produce \emph{time-variant} geometry~\cite{sucar2025vdpm} without dense correspondences across time.
MotionCrafter~\cite{zhu2026motioncrafter} extends Geo4D to predict scene flow but only between neighboring frames, and MoMaps~\cite{lei2025momaps} generates per-pixel 3D trajectories (our $\mathcal{R}$) from a single image with an image diffusion model.
The concurrent TrackCraft3R~\cite{nam2026trackcraft3r} adapts a video DiT with LoRA to predict reference-anchored 3D track residuals; it retains the output VAE, whereas our regression readout removes it (\cref{sec:regression,sec:regression-results}).

A second line of work observes that VDMs already \emph{contain} 2D tracking signal in their internal features:
DIFT~\cite{tang2023dift} extracts semantic correspondence from image diffusion features;
DiffTrack~\cite{nam2025difftrack} extends this to video DiTs;
Track4Gen~\cite{jeong2025track4gen} uses these features as an auxiliary signal to improve generated-video consistency;
and subsequent work~\cite{shrivastava2026pointprompting,son2025ditracker} achieves competitive sparse 2D tracking by reading from frozen VDM features.
\nickname{} differs by \emph{generating} dense 3D tracking videos as model output, rather than reading sparse 2D tracks from frozen features --- giving, to our knowledge, an early systematic study of how to represent dense 3D tracking inside a pretrained VDM\@.

\section{Method}%
\label{sec:method}

Given a monocular video $\mathcal{V} \in \mathbb{R}^{N \times H \times W \times 3}$, \nickname{} recovers dense 3D point trajectories in a fixed reference frame.
We assume that camera parameters and depth are available from an off-the-shelf reconstructor and focus on the tracking problem.

\begin{figure*}[t]
\centering
\begin{minipage}[c]{0.35\textwidth}
\centering
\includegraphics[width=\linewidth]{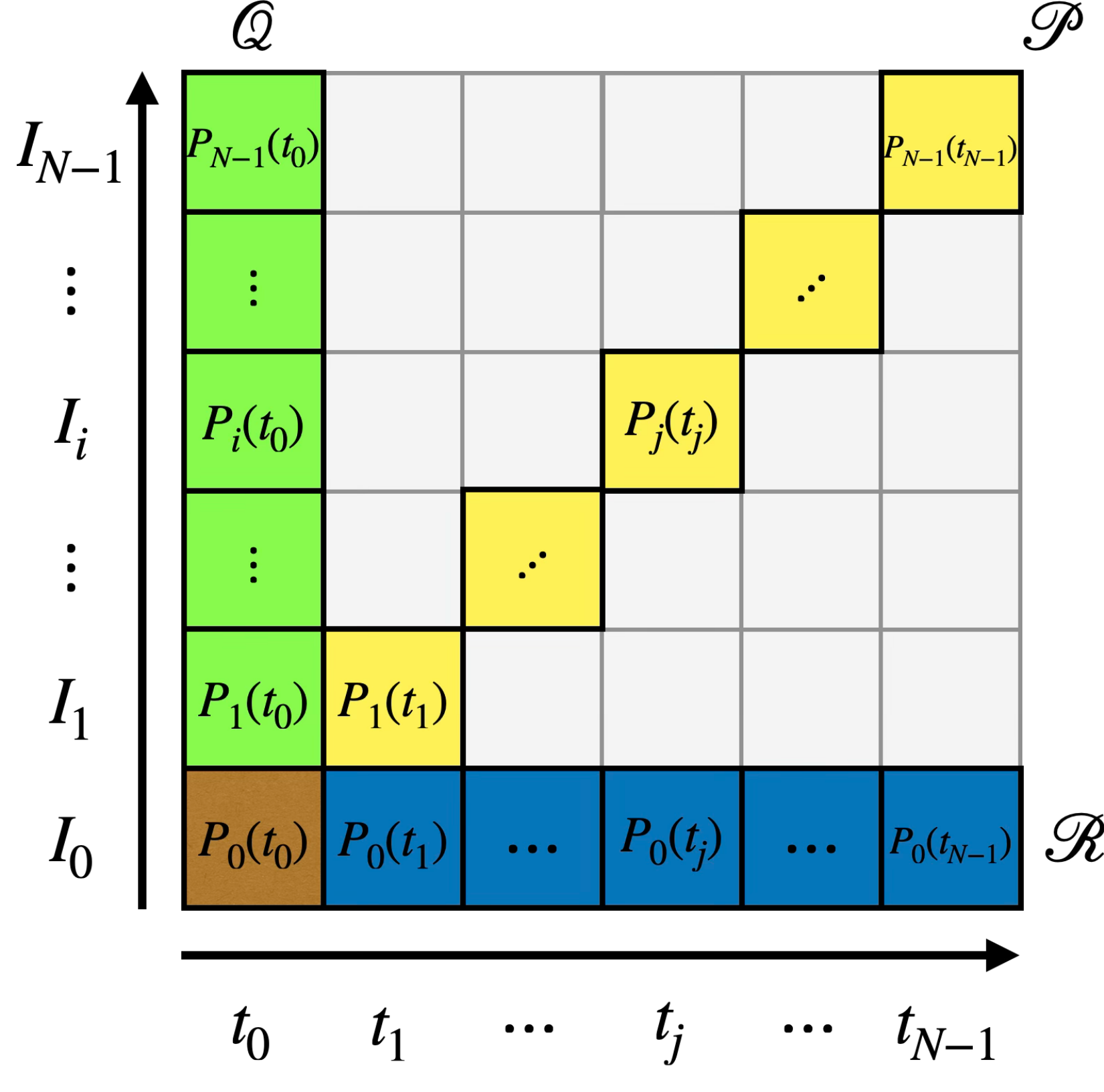}
\end{minipage}\hfill
\begin{minipage}[c]{0.6\textwidth}
\caption{\textbf{Prediction formulations} (adapted from V-DPM~\cite{sucar2025vdpm})\textbf{.}
Each entry $\pzero{i}{j}$ (short for $P_i(t_j,\pi_0)$, with the reference frame $\pi_0$ fixed) gives the positions at time $t_j$ of the surface points observed in frame $I_i$.
$\mathcal{P}$ (diagonal): geometry at each frame's capture time.
$\mathcal{Q}$ (column): time-invariant prediction~\cite{sucar2025vdpm}, referring every frame $I_i$ back to the reference time $t_0$.
$\mathcal{R}$ (row): dense 3D tracking, propagating the query-frame point map $P_0$ across all timestamps.
\nickname{} encodes $\mathcal{R}$ explicitly as 3D offsets (\rtv) and $\mathcal{Q}$ as normalized canonical colors (\ncmv), from which $\mathcal{R}$ is recovered by 2D tracking.
}
\label{fig:vdpm_pointmap_vis}
\end{minipage}
\end{figure*}

\subsection{Video-like representations of dynamics}%
\label{sec:preliminary}

Let $\mathcal{V} = \{I_i\}_{i=0}^{N-1}$ be a sequence of $N$ RGB frames captured at times $t_i$ from cameras with extrinsics $\pi_i \in \mathbb{SE}(3)$, and $u$ a pixel location.
A \emph{Dynamic Point Map}~\cite{sucar2025vdpm} $P_i(t,\pi)(u)$ assigns to each pixel $u$ in $I_i$ its 3D location at time $t$ expressed in reference frame $\pi$.
We fix $\pi = \pi_0$ and abbreviate $P_i(t)$ for $P_i(t,\pi_0)$, so 3D coordinates are viewpoint-invariant.

As $i,j$ vary from $0$ to $N-1$, the point maps $P_i(t_j)$ form a 2D grid (\cref{fig:vdpm_pointmap_vis}).
The diagonal
\begin{equation}%
\label{eq:pixel-aligned-pmap}
\mathcal{P} = \big( \pzero{0}{0},\; \pzero{1}{1},\; \ldots,\; \pzero{N-1}{N-1} \big)
\end{equation}
records the 3D position of pixels at their capture time and is recoverable from depth and camera parameters alone.
Tracking information lives in the off-diagonal entries:
\begin{equation}
\label{eq:def:vdpm-col}
\mathcal{Q} = \big( \pzero{0}{0},\; \pzero{1}{0},\; \ldots,\; \pzero{N-1}{0} \big),
\quad
\mathcal{R} = \big( \pzero{0}{0},\; \pzero{0}{1},\; \ldots,\; \pzero{0}{N-1} \big).
\end{equation}
$\mathcal{Q}$ is a collection of \emph{time-invariant} point maps: the 3D coordinates of each pixel referred back to the fixed reference time $t_0$, undoing scene motion (the V-DPM representation~\cite{sucar2025vdpm,sucar25dynamic}). As a video, $\mathcal{Q}$ is spatially aligned to $\mathcal{V}$ frame-by-frame.
$\mathcal{R}$ is dual: a collection of \emph{time-dependent} point maps that forward pixels of the reference frame $I_0$ to their 3D location at later times $t_j$ --- as a video, this is spatially aligned with $I_0$ but color-varying~\cite{cotracker,karhade2025any4d,harley2025alltracker}; MoMaps~\cite{lei2025momaps} calls this image of per-pixel 3D trajectories a \emph{motion map} and generates it from a single image.

\begin{figure}[t]
\centering
\includegraphics[width=\textwidth]{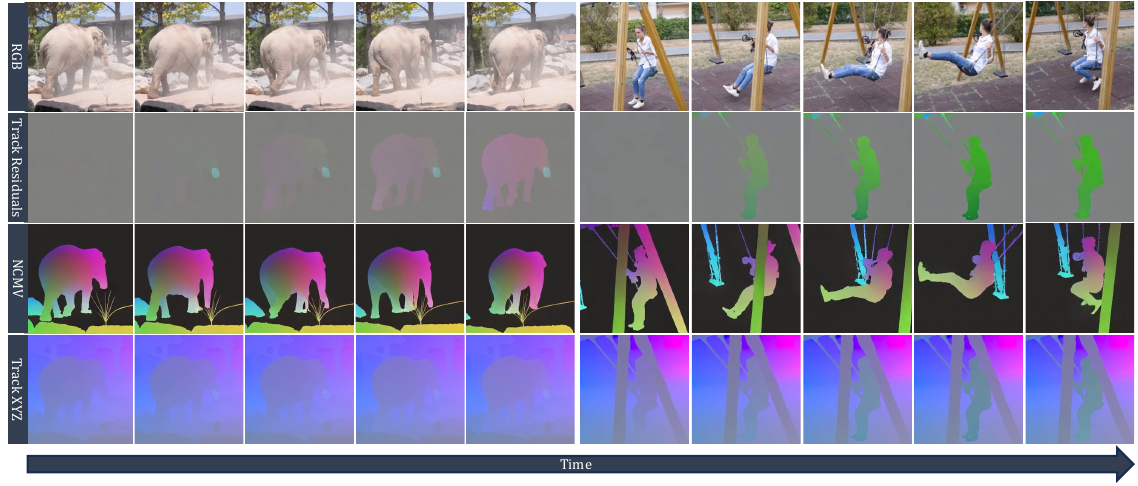}
\caption{\textbf{Two video representations of dense 3D tracking on DAVIS~\cite{davis_CVPR_2016}.}
Rows show RGB, \rtv{} offsets, \ncmv{} canonical coordinates, and recovered XYZ tracks.
Adding inverse-normalized \rtv{} offsets to the query-frame point map recovers the tracks. All predictions are \nickname{} outputs.}%
\label{fig:two-rep=for-vdm}
\end{figure}
\subsection{\nickname}%
\label{sec:ours}

Given $\mathcal{V}$ and the diagonal point maps $\mathcal{P}$ from an off-the-shelf 3D reconstructor (geometric scaffold), we sample row $\mathcal{R}$ and column $\mathcal{Q}$ from the conditional distribution
\begin{equation}
\mathbb{P} \Bigl(
    \mathcal{R}, \mathcal{Q} \;\Big|\; \mathcal{V}, \mathcal{P} \Bigr),
\end{equation}
using a pretrained video diffusion model (VDM). $\mathcal{R}$ encodes full 3D tracks directly; $\mathcal{Q}$ does so indirectly but is spatially aligned with $\mathcal{V}$ and resembles natural video.

Point-map values lie outside the $[-1,1]$ range expected by the video VAE\@.
A naïve global normalization to $[-1,1]$~\cite{geo4d,chen20254dnex} works for $\mathcal{Q}$ but wastes dynamic range for $\mathcal{R}$, since dynamic regions occupy only a small portion of the scene.
Following~\cite{zhu2026motioncrafter} we instead represent $\mathcal{R}$ as a 3D offset:
\begin{equation}
\label{eq:track_offset}
\pzero{i}{j} = \pzero{i}{0} + \bm{\Delta}_{i}^{j},
\end{equation}
where $\bm{\Delta}_{i}^{j}$ denotes the deformation vector ($\Delta x, \Delta y, \Delta z$) of the point map $\pzero{i}{j}$ relative to the query frame $\pzero{i}{0}$ (the resulting tracks are in \cref{fig:two-rep=for-vdm}, row 4).
We define this offset across arbitrary timestamps, not just neighboring ones as in~\cite{zhu2026motioncrafter}.
To further improve sensitivity to small displacements, we predict offsets in log space~\cite{lin2026depth,xu2025pixel}:
\begin{equation}
\tilde{\bm{\Delta}}_{i}^{j} = \mathrm{sign}(\bm{\Delta}_{i}^{j}) \cdot \log \! \bigl(|\bm{\Delta}_{i}^{j}| + 1\bigr) \cdot K.
\label{eq:track_offset_normalization}
\end{equation}
With this transform, static regions with zero deformation remain exactly zero.
We choose a scalar $K$ so that $\tilde{\bm{\Delta}}_{i}^{j}\in[-1,1]$, where $K$ is computed from the training set statistics.
We thus encode $\tilde{\bm{\Delta}}$, which we call Residual 3D Tracking Video (\rtv) (\cref{fig:two-rep=for-vdm}, row 2), with our autoencoder in lieu of $\mathcal{R}$.

As for $\mathcal{Q}$, we map it to the range $[-1,1]$ linearly, and call the output a Normalized Coordinate Map Video (\ncmv)~\cite{Wang_2019_NOCS,wang2024crm} (\cref{fig:two-rep=for-vdm}, row 3).
Specifically, we define a normalization function $\phi: \mathbb{R}^3 \to [-1,1]^{3}$ that maps the 3D coordinates of the foreground objects to the unit cube:
\begin{equation}
\phi(\mathbf{p}) = 2\,\bar{\mathbf{p}} - \mathbf{1}, \quad \text{where} \quad
\bar{\mathbf{p}} = \frac{\mathbf{p} - \mathbf{p}_{\min}}{\mathbf{p}_{\max} - \mathbf{p}_{\min}}.
\label{eq:ncm_norm}
\end{equation}
Here $\mathbf{p} \in \pzero{0}{0}$ is a 3D point extracted from the point map $\pzero{0}{0}$ for the query frame at $t=0$, and $\mathbf{p}_{\min}$ and $\mathbf{p}_{\max}$ are the computed minimum and maximum values from the point map, respectively.
Using $\phi$, we define the \ncmv{} to be $M = \Phi(\mathcal{Q})$, where $\Phi$ is applied component-wise.
Intuitively, this amounts to `colorizing' each pixel of the reference image $I_0$ with a normalized version of the corresponding physical point coordinates
\begin{equation}
\label{eq:ncm_query}
M_0(u) = \phi \bigl(\pzero{0}{0}(u)\bigr).
\end{equation}
The value $M_0(u)$ can also be thought of as a unique identifier assigned to that pixel.
Subsequent frames $j > 0$ are colorized in a similar manner.
The end result is that each pixel $u'$ at time $t_j$ receives the same 3D point/color/identifier assigned to its counterpart $u$ at time $t_0$, where the correspondence $(u,u')$ is given by the 3D tracking:
\begin{equation}
M_j(u') = M_0(u), \quad \text{where } u' \text{ is the projection of } \pzero{0}{j}(u) \text{ into camera } \pi_j.
\label{eq:ncm_prop}
\end{equation}
By construction, pixels corresponding to the same 3D surface point share a consistent and unique color throughout the video sequence.
We expect this representation to be easier for the VDM to learn, consistent with recent findings that generative models implicitly capture dense correspondences~\cite{lan2022ddf_ijcv,mu2022coordgan,tang2023dift}.

\paragraph{Recovering 3D tracking from \ncmv.}

In principle, we can decode \ncmv{} colors into tracks by matching points by their unique canonical `color' across frames, but this performs poorly due to the finite distinctiveness and accuracy of the signal (\cref{sec:ablation}).
Instead, we run CoTracker3~\cite{cotracker3} on $M$ to obtain dense 2D correspondences and lift them to 3D using VGGT~\cite{wang25vggt} on the original video $\mathcal{V}$.

\paragraph{Video diffusion model.}

We adopt Wan~\cite{wan2025} as our backbone: its 3D VAE encodes videos into spatio-temporal latents and the DiT~\cite{Peebles2022DiT} applies full attention over all latent tokens, in contrast to factorized spatial-temporal attention~\cite{xing2023dynamicrafter}.
Text conditioning uses umT5~\cite{chung2023unimax}.

Both \rtv{} and \ncmv{} are $3$-channel videos and use the unmodified Wan VAE\@; we add no new output heads. Adaptation uses LoRA~\cite{hu2022lora} on the DiT attention/MLP linears (rank $64$), with VAE and patch embedder frozen --- see \cref{appendix:arch} for full layer-level details.

\subsection{{Generating multimodal 3D tracking}}%
\label{sec:model}

\nickname{} jointly generates \rtv{} and \ncmv{} in one iterative sampling process, which lets us compare the representations head-to-head within a shared architecture. \rtv{} is the output we recommend for deployment; \ncmv{} is retained as an auxiliary representation and training objective.

We extend an I2V VDM to jointly generate two 3-channel videos, $\{\tilde{\bm{\Delta}}_{0}^{j}\}_{j=0}^{N-1}$ (\rtv) and $\{M_j\}_{j=0}^{N-1}$ (\ncmv), conditioned on $\mathcal{V}$ and $\mathcal{P}$.
All modalities are first encoded into the latent space via the VAE encoder $\mathcal{E}(\cdot)$, and concatenated along the width dimension to form unified latent representations:
\begin{equation}
\mathcal{Z}_1 = [x_{\tilde{\bm{\Delta}}};\; x_{M}], \quad \mathcal{C} = [x_{\mathcal{V}};\; x_{\mathcal{P}}], \quad \mathcal{Z}_1, \mathcal{C} \in \mathbb{R}^{n\times h \times 2w \times c},
\end{equation}
where $x_{*} = \mathcal{E}(*)$, $\mathcal{Z}_1$ denotes the clean generation target, $\mathcal{C}$ denotes the conditioning latents (kept clean throughout training), $[\cdot \, ; \, \cdot]$ indicates concatenation along the width dimension~\cite{chen20254dnex,huang2026gen3r3dscenegeneration}, and $n,h,w,c$ denote the spatio-temporal dimensions after the video VAE $\mathcal{E}$.
We adopt width-wise concatenation to allow the model to flexibly predict non-pixel-aligned outputs~(\cref{fig:two-rep=for-vdm}, row 2), whereas channel-wise concatenation is better suited to input-output aligned predictions~\cite{vace,zhang2023adding,li2025egoedit}.

Training follows Rectified Flow~\cite{lipman2022,liu2022flow}: with $\mathcal{Z}_0 \sim \mathcal{N}(0,I)$, $t \sim \mathcal{U}(0,1)$, and $\mathcal{Z}_t = t\mathcal{Z}_1 + (1-t)\mathcal{Z}_0$, we regress the velocity $v_t = \mathcal{Z}_1 - \mathcal{Z}_0$:
\begin{equation}
\mathcal{L}_{\mathrm{FM}} =
\mathbb{E}_{\mathcal{Z}_0,\mathcal{Z}_1,\mathcal{C},t}
\bigl \|
u_{\bm{\theta}}([\mathcal{Z}_t;\mathcal{C}],t) - v_t
\bigr \|_2^2,
\quad [\mathcal{Z}_t;\mathcal{C}] \in \mathbb{R}^{n \times h \times 4w \times c}.
\end{equation}
Four learnable domain embeddings differentiate the modalities, following~\cite{chen20254dnex}.

\subsection{Regression readout without the output VAE}%
\label{sec:regression}

The generative readout decodes tracking through the video VAE, which blurs depth discontinuities (\cref{appendix:artifact-analysis}) and needs a target at every pixel.
Our third readout keeps the VDM prior and the residual target of \cref{eq:track_offset} but \emph{regresses} the residual from the VDM features: without a VAE on the output path, the loss applies only at annotated pixels, so sparse tracking datasets can be used.
Following TrackCraft3R~\cite{nam2026trackcraft3r}, the model is anchored at the query frame: for every pixel $u$ of $I_0$ and time $t_j$ it predicts a normalized residual $\bm{\delta}_j(u)$ and composes
\begin{equation}
\label{eq:regression}
\hat{P}_0(t_j)(u) = \pzero{0}{0}(u) + r\,\bm{\delta}_j(u),
\end{equation}
with $r$ the per-clip normalization radius of $\mathcal{P}$.
In \cref{fig:method_overview}b, the point maps enter through a small patch encoder rather than the VAE, so every frame is the pair $[x_{\mathcal{V}};\, g_{\mathcal{P}}]$ of its RGB latent and geometry features.
The LoRA-adapted Wan2.2-TI2V-5B~\cite{wan2025} \emph{world expert}~\cite{yuan2026fastwam} and a private \emph{motion expert} of equal depth process all-frame and repeated query-frame tokens, respectively, with joint attention at every layer~\cite{liang2024mixture}.
A DPT head~\cite{ranftl21vision,wang26vggt-omega} on four motion-expert layers outputs the residual and a visibility logit in a single forward pass, without noise or denoising.
TrackCraft3R instead retains the output VAE. Architecture and training details are in \cref{appendix:regression,fig:supp-mot-arch}.

\section{Experiments}%
\label{sec:experiments}

\subsection{Settings}\label{sec:exp-settings}

\paragraph{Datasets.}

In its generative form, \nickname{} requires video sequences with \emph{dense} 3D tracking annotations, which are scarce in the real world, so we train only on synthetic data; the regression readout (\cref{sec:regression}, \cref{appendix:direct-readout}) relaxes this to sparse supervision.
We re-render Kubric~\cite{greff2021kubric} into $5{,}600$ clips of $49$ frames with dense 3D tracks computed following Delta3D~\cite{ngo2024delta}.
We additionally include TartanAir~\cite{tartanair2020iros}, a high-quality synthetic dataset with negligible object motion, with tracking ground truth set to the static query-frame point map.
We evaluate on five benchmarks:
PointOdyssey (PO)~\cite{zheng2023point} (synthetic, complex motion);
Aria Digital Twin (ADT)~\cite{pan2023ariadigitaltwinnew} (real-world egocentric with ground-truth 3D);
Panoptic Studio (PStudio)~\cite{pstudio} (real-world multi-person scenes);
Dynamic Replica (DR)~\cite{karaev2023dynamicstereo} (synthetic indoor with dynamic objects);
and a held-out Kubric test split disjoint from training.
We use $50$ sequences per benchmark and their first $49$ frames ($49$ ADT sequences in the regression replays, \cref{appendix:regression}).
The first four have only \emph{sparse} annotations. Our generative models are zero-shot on all four, unlike prior models~\cite{st4rtrack2025,liu2025traceanythingrepresentingvideo,karhade2025any4d,sucar2025vdpm} trained on one or more, and \nickname-Reg, trained on PointOdyssey and Dynamic Replica (\cref{sec:regression-results}).

\paragraph{Metrics.}

Following St4RTrack~\cite{st4rtrack2025} and TAPVid-3D~\cite{koppula24tapvid-3d:}, we report \textit{Average Percent of Points within Delta} ($\text{APD}_{\text{3D}}$) at $\delta_{\text{3D}} \in \{0.1, 0.3, 0.5, 1.0\}$~m over $49$ frames, and \textit{End-Point Error} (EPE) after a global $\operatorname{Sim}(3)$ alignment of the predicted tracks to the ground truth~\cite{umeyama91least-squares}, as in 4RC~\cite{luo264rc:}; all results in \cref{tab:track_result} are $\operatorname{Sim}(3)$ aligned.
The generative readouts do not predict visibility; we report APD/EPE for all methods, including \nickname-Reg.

\paragraph{Baselines and comparison protocol.}

We compare \nickname{} against four groups of methods that produce 3D trajectories:
\textbf{(i)} 2D-tracker pipelines that lift 2D tracks to 3D via a depth backbone --- SpatialTrackerV2~\cite{xiao2025spatialtracker} and CoTracker3$^*$~\cite{cotracker3} (lifted by VGGT~\cite{wang25vggt});
\textbf{(ii)} dense 4D reconstruction methods --- St4RTrack~\cite{st4rtrack2025} and TraceAnything~\cite{liu2025traceanythingrepresentingvideo};
\textbf{(iii)} concurrent methods built on stronger 3D foundation backbones --- Any4D~\cite{karhade2025any4d} and V-DPM~\cite{sucar2025vdpm}; and \textbf{(iv)} TrackCraft3R~\cite{nam2026trackcraft3r}, replayed with the same STream3R geometry.
DELTA~\cite{ngo2024delta,ngo2025deltav2} and TAPIP3D~\cite{tapip3d} use incompatible conditioning interfaces (no point-map input) and are excluded from this comparison.
Each baseline runs in its native protocol with its own depth/camera estimator as released; \nickname{} uses STream3R~\cite{stream3r2025} or DAv3~\cite{lin2026depth} as the scaffold, all evaluated under the D4RT~\cite{zhang2025d4rt} protocol; this is a \emph{protocol} comparison rather than an isolation of the conditioning factor.

\paragraph{Implementation and inference cost.}
The joint model of \cref{tab:track_result} trains a rank-$64$ LoRA~\cite{hu2022lora} on the official Wan~\cite{wan2025} codebase, continuing from an RTV-only adapter for $5{,}000$ iterations on $32$ GH200 GPUs (about one day; \cref{appendix:arch}).

Inference uses $50$ denoising steps with random latent noise ($0$--$400$ steps) on the input video latent~\cite{bai25recammaster:} to reduce the synthetic-to-real gap. In the 5B sampling ablation, $50$, $10$ and $5$ steps take $14$, $9.6$ and $8.6$\,s per clip on one A6000 with similar EPE (\cref{appendix:sampling_steps}).

\subsection{Qualitative evaluation}
\begin{figure}[t]
\centering
\includegraphics[width=\textwidth]{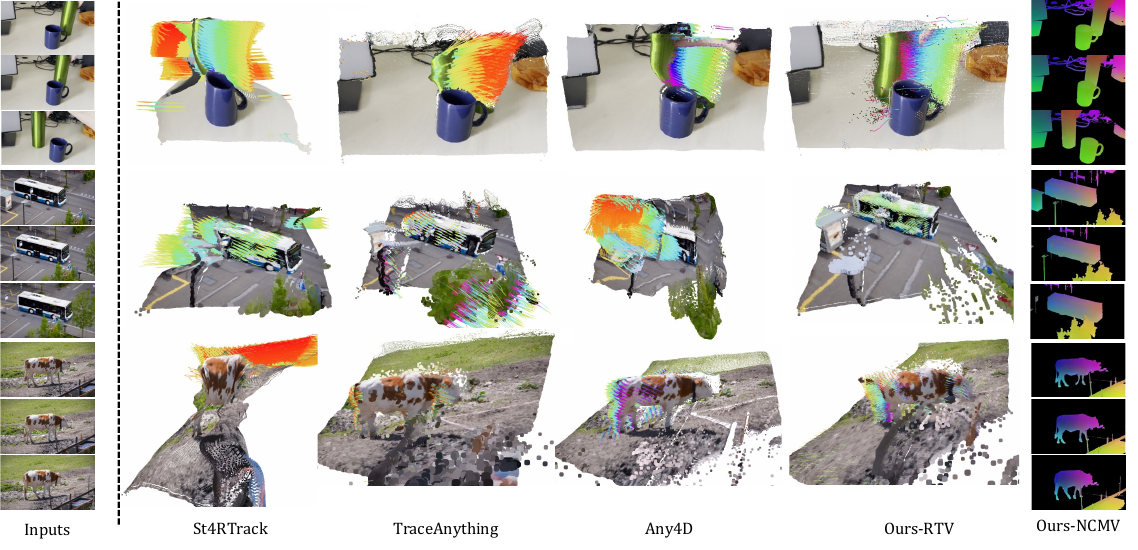}
\caption{\textbf{Qualitative comparison of dynamic tracking on DAVIS~\cite{davis_CVPR_2016}.}
Dense 3D tracking is visualized as colored point clouds, with color encoding query-frame identity of each tracked point.
Left: input frames. Right: reconstructions from St4RTrack~\cite{st4rtrack2025}, TraceAnything~\cite{liu2025traceanythingrepresentingvideo}, Any4D~\cite{karhade2025any4d}, and \nickname{} (ours).
\nickname{} produces more temporally coherent trajectories with fewer outliers, particularly on articulated and deformable motion (\eg moving mug, walking cows).
All baselines are run from their official released checkpoints and inference scripts at the resolution recommended by the authors;
sequences shown were selected to expose dynamic-object failure modes for the comparators.
More DAVIS results appear in \cref{fig:supp-more-results}.}%
\label{fig:main-comparison}
\end{figure}
\Cref{fig:main-comparison} visualizes zero-shot 3D reconstruction and tracking on DAVIS~\cite{davis_CVPR_2016}; conditioning point maps $\mathcal{P}$ come from STream3R~\cite{stream3r2025} or DAv3~\cite{lin2026depth}.
\nickname{} produces temporally coherent dense trajectories that follow camera-induced parallax and dynamic object motion. Selected cases highlight dynamic-object failures; broader DAVIS results appear in \cref{appendix:more}, and \cref{fig:davis-reg} compares all three readouts with TrackCraft3R.

\subsection{Quantitative evaluation}%
\label{sec:quantitative}

\begin{table*}[!t]
\centering
\captionsetup{font=footnotesize}
\caption{\textbf{Sparse 3D point tracking ($\operatorname{Sim}(3)$ aligned).} APD$\uparrow$ and EPE$\downarrow$ after a global $\operatorname{Sim}(3)$ alignment to the ground truth, as in 4RC~\cite{luo264rc:}, on all points (All) and dynamic points (Dyn). Bold and underline mark \textbf{best} and \underline{second-best} results. CoTracker$^*$: CoTracker lifted by VGGT\@. \nickname-NCMV and \nickname-RTV are the two readouts of the joint Wan2.1-I2V-14B model; \nickname-NCMV predicts dynamic points only. $^\ddagger$Official checkpoint replayed with STream3R geometry under the same scorer (dynamic ADT not reported). $^\dagger$\nickname-Reg (\cref{sec:regression}): Wan2.2-5B, STream3R scaffold, eight training sources including PointOdyssey and Dynamic Replica.}%
\label{tab:track_result}
\fontsize{7.5}{9}\selectfont
\renewcommand{\arraystretch}{1.05}
\setlength{\tabcolsep}{1pt}
\begin{tabular*}{\linewidth}{@{\extracolsep{\fill}}lcccccccccccccc@{}}
\toprule
& \multicolumn{4}{c}{\textbf{PointOdyssey}~\cite{zheng2023point}}
& \multicolumn{4}{c}{\textbf{ADT}~\cite{pan2023ariadigitaltwinnew}}
& \multicolumn{2}{c}{\textbf{PStudio}~\cite{pstudio}}
& \multicolumn{4}{c}{\textbf{Dynamic Replica}~\cite{karaev2023dynamicstereo}} \\
\cmidrule(lr){2-5}\cmidrule(lr){6-9}\cmidrule(lr){10-11}\cmidrule(l){12-15}
& \multicolumn{2}{c}{All} & \multicolumn{2}{c}{Dyn} & \multicolumn{2}{c}{All} & \multicolumn{2}{c}{Dyn} & \multicolumn{2}{c}{Dyn} & \multicolumn{2}{c}{All} & \multicolumn{2}{c}{Dyn} \\
\cmidrule(lr){2-3}\cmidrule(lr){4-5}\cmidrule(lr){6-7}\cmidrule(lr){8-9}\cmidrule(lr){10-11}\cmidrule(lr){12-13}\cmidrule(l){14-15}
\textbf{Method} & APD$\uparrow$ & EPE$\downarrow$ & APD$\uparrow$ & EPE$\downarrow$ & APD$\uparrow$ & EPE$\downarrow$ & APD$\uparrow$ & EPE$\downarrow$ & APD$\uparrow$ & EPE$\downarrow$ & APD$\uparrow$ & EPE$\downarrow$ & APD$\uparrow$ & EPE$\downarrow$ \\
\midrule
CoTracker$^*$~\cite{cotracker3} & 0.6165 & 0.4721 & 0.4722 & 0.5264 & 0.8303 & 0.1664 & 0.9157 & \textbf{0.0397} & 0.7579 & 0.2031 & 0.6319 & 0.3842 & 0.6456 & 0.3014 \\
SpaTrackerV2~\cite{xiao2025spatialtracker} & 0.2655 & 1.1016 & 0.3512 & 0.6921 & 0.5594 & 0.5158 & 0.8509 & 0.0621 & 0.5325 & 0.4211 & 0.6264 & 0.4052 & 0.5380 & 0.3910 \\
St4RTrack~\cite{st4rtrack2025} & 0.6751 & 0.3126 & 0.7380 & \underline{0.2146} & 0.7848 & 0.1990 & 0.8864 & 0.0526 & 0.8002 & 0.1709 & 0.7353 & 0.2651 & \underline{0.7944} & \underline{0.1681} \\
TraceAnything~\cite{liu2025traceanythingrepresentingvideo} & 0.2629 & 1.1003 & 0.4323 & 0.5798 & 0.8156 & 0.1773 & 0.8708 & 0.0559 & 0.6469 & 0.3085 & 0.3766 & 0.7415 & 0.5418 & 0.3972 \\
\midrule
Any4D~\cite{karhade2025any4d} & 0.5332 & 0.6213 & 0.4012 & 0.6243 & 0.7955 & 0.1787 & 0.8633 & 0.0592 & 0.7702 & 0.1971 & 0.7484 & 0.2647 & 0.6668 & 0.2723 \\
V-DPM~\cite{sucar2025vdpm} & \textbf{0.8350} & \textbf{0.1726} & \textbf{0.8169} & \textbf{0.1688} & \underline{0.8880} & \textbf{0.1109} & 0.8590 & 0.0593 & \textbf{0.9180} & \textbf{0.0916} & \textbf{0.8065} & \textbf{0.2032} & 0.7761 & 0.1900 \\
TrackCraft3R$^\ddagger$~\cite{nam2026trackcraft3r} & 0.7904 & 0.2227 & \underline{0.7734} & 0.2494 & 0.8874 & 0.1163 & {--} & {--} & 0.8418 & 0.1400 & 0.7839 & 0.2289 & 0.7571 & 0.2207 \\
\midrule
\rowcolor{gray!10}[0pt][0pt]\oursrowshade {\nickname}-NCMV & {--} & {--} & 0.4798 & 0.5162 & {--} & {--} & 0.8855 & \underline{0.0513} & 0.7495 & 0.2098 & {--} & {--} & 0.6587 & 0.2873 \\
\rowcolor{gray!10}[0pt][0pt]\oursrowshade \nickname-RTV & 0.7786 & 0.2289 & 0.6896 & 0.2887 & \textbf{0.8920} & \underline{0.1130} & \textbf{0.9541} & 0.0610 & 0.8101 & 0.1596 & 0.7751 & 0.2374 & 0.7847 & 0.1859 \\
\rowcolor{gray!10}[0pt][0pt]\oursrowshade \nickname-Reg$^\dagger$ & \underline{0.8050} & \underline{0.2078} & 0.7155 & 0.2691 & 0.8854 & 0.1168 & \underline{0.9430} & 0.0675 & \underline{0.8572} & \underline{0.1311} & \underline{0.7871} & \underline{0.2284} & \textbf{0.8141} & \textbf{0.1674} \\
\bottomrule
\end{tabular*}
\end{table*}

\paragraph{Results overview.}

\Cref{tab:track_result} reports 3D tracking on the four benchmarks with sparse annotations for \textbf{(a)} \rtv{} (\nickname-RTV) and \textbf{(b)} \ncmv{} (\nickname-NCMV) from the joint model, and \textbf{(c)} the regression readout \nickname-Reg (\cref{sec:regression-results}) with the same scaffold.

\paragraph{Dense Kubric.}

On the held-out dense Kubric split (\cref{tab:kubric}) --- the only benchmark as dense as our training signal --- \nickname-RTV ($0.336$ APD / $1.079$ EPE) outperforms St4RTrack ($0.047$ / $5.333$) and TraceAnything ($0.288$ / $1.550$) by a clear margin; on dynamic points V-DPM is best.

\paragraph{Comparison with feed-forward 3D tracking.}
SpatialTrackerV2~\cite{xiao2025spatialtracker}, St4RTrack~\cite{st4rtrack2025}, Any4D~\cite{karhade2025any4d} and V-DPM~\cite{sucar2025vdpm} are trained on splits of one or more evaluation datasets (\eg, PointOdyssey, Dynamic Replica), whereas the generative \nickname{} has dynamic tracking labels only from Kubric and is zero-shot on the four external benchmarks.
\nickname-RTV is nevertheless competitive: on ADT it ranks first in APD on both All ($0.892$) and Dyn ($0.954$), and on PStudio and DR it is comparable to or better than St4RTrack and Any4D.
V-DPM is strongest on PO, PStudio and DR All; on PO it is trained on the same benchmark and builds on a VGGT backbone trained on broader data (\cref{appendix:data-efficiency}).

\paragraph{\rtv{} vs.\ \ncmv: representation comparison.}
On dynamic points (where \ncmv{} produces predictions), \nickname-\rtv{} consistently outperforms \nickname-\ncmv{}, \eg, $0.690$ vs $0.480$ APD on PO Dyn: \rtv{} reads 3D offsets directly (\cref{eq:track_offset}), while \ncmv{} adds a 2D-tracker decoding stage that compounds errors.
Yet \nickname-\ncmv{} \emph{still} outperforms CoTracker3$^*$~\cite{cotracker3} (same VGGT lifting) on PO Dyn ($0.480$ vs $0.472$) and DR Dyn ($0.659$ vs $0.646$), and is comparable on PStudio and ADT: feeding CoTracker the VDM-generated \ncmv{} instead of RGB improves APD on PO and DR, evidence that a VDM can produce a structured dense correspondence signal as output, not only as an internal feature~\cite{tang2023dift,nam2025difftrack,jeong2025track4gen}.
We therefore recommend \rtv{} for deployment and treat \ncmv{} as a probe of VDM correspondence priors and an auxiliary training target (\cref{appendix:ncmv-analysis}).

\subsection{Ablations and analysis}%
\label{sec:ablation}
\definecolor{impStrong}{RGB}{170, 215, 170}
\definecolor{impMild}{RGB}{210, 235, 210}
\definecolor{degStrong}{RGB}{245, 170, 170}
\definecolor{degMild}{RGB}{250, 210, 210}

\begin{table*}[htbp]
\centering
\captionsetup{font=footnotesize}
\caption{\textbf{Ablations of design choices.}
Baseline is \nickname{} (Wan 2.2 5B, STream3R).
Top three rows modify one component each;
bottom rows scale the VDM backbone (1.3B / 14B) and swap conditioning (DAv3 / STream3R).
Green/red show gains/losses over baseline; intensity indicates magnitude.
All models are trained and evaluated at $256{\times}256$ following~\cite{zhang2025d4rt}.
Reported on ADT~\cite{pan2023ariadigitaltwinnew}, PStudio~\cite{pstudio}, and DR~\cite{karaev2023dynamicstereo};
PO~\cite{zheng2023point} is excluded due to mismatched aspect ratios.
}%
\label{tab:ablation_progressive}
\renewcommand{\arraystretch}{1.05}
\setlength{\tabcolsep}{2pt}
\fontsize{7.5}{9}\selectfont
\begin{tabular*}{\linewidth}{@{\extracolsep{\fill}}l cccc cc cccc@{}}
\toprule
\multirow{3}{*}{\textbf{Variant}}
& \multicolumn{4}{c}{\textbf{ADT}~\cite{pan2023ariadigitaltwinnew}}
& \multicolumn{2}{c}{\textbf{PStudio}~\cite{pstudio}}
& \multicolumn{4}{c}{\textbf{DR}~\cite{karaev2023dynamicstereo}} \\
\cmidrule(lr){2-5} \cmidrule(lr){6-7} \cmidrule(lr){8-11}
& \multicolumn{2}{c}{All}
& \multicolumn{2}{c}{Dyn}
& \multicolumn{2}{c}{Dyn}
& \multicolumn{2}{c}{All}
& \multicolumn{2}{c}{Dyn} \\
\cmidrule(lr){2-3} \cmidrule(lr){4-5} \cmidrule(lr){6-7} \cmidrule(lr){8-9} \cmidrule(lr){10-11}
& APD$\uparrow$ & EPE$\downarrow$
& APD$\uparrow$ & EPE$\downarrow$
& APD$\uparrow$ & EPE$\downarrow$
& APD$\uparrow$ & EPE$\downarrow$
& APD$\uparrow$ & EPE$\downarrow$ \\
\midrule
{\nickname (Wan 2.2 5B)}
& 0.8831 & 0.1196
& 0.9370 & 0.0722
& 0.7281 & 0.2231
& 0.5245 & 0.4468
& 0.7009 & 0.2536 \\
\midrule
w/o Residual Pred
& \cellcolor{degStrong}0.6625 & \cellcolor{degStrong}0.3235
& \cellcolor{degMild}0.9218   & \cellcolor{degMild}0.0859
& \cellcolor{degStrong}0.6100 & \cellcolor{degStrong}0.3358
& \cellcolor{degStrong}0.2959 & \cellcolor{degStrong}0.8439
& \cellcolor{degStrong}0.4837 & \cellcolor{degStrong}0.4756 \\
w/o Log Norm
& \cellcolor{degMild}0.8793 & \cellcolor{degMild}0.1217
& \cellcolor{degMild}0.9200   & \cellcolor{degMild}0.0837
& \cellcolor{degMild}0.7109 & \cellcolor{degMild}0.2384
& \cellcolor{degMild}0.5109 & \cellcolor{degMild}0.4668
& \cellcolor{degMild}0.6857 & \cellcolor{degMild}0.2690 \\
w/o XYZ Cond
& \cellcolor{degMild}0.8722   & \cellcolor{degMild}0.1289
& \cellcolor{degStrong}0.8974 & \cellcolor{degStrong}0.1046
& \cellcolor{degStrong}0.6396 & \cellcolor{degStrong}0.3106
& \cellcolor{degStrong}0.4386 & \cellcolor{degStrong}0.5718
& \cellcolor{degStrong}0.5908 & \cellcolor{degStrong}0.3608 \\
\midrule
{\nickname (Wan 2.1 1.3B)}
& \cellcolor{degMild}0.8588   & \cellcolor{degMild}0.1391
& \cellcolor{degStrong}0.8601 & \cellcolor{degStrong}0.1295
& \cellcolor{degStrong}0.6928 & \cellcolor{degStrong}0.2574
& \cellcolor{impMild}0.5283   & \cellcolor{impMild}0.4425
& \cellcolor{degMild}0.6999   & \cellcolor{degMild}0.2549 \\
{\nickname (Wan 2.1 14B, DAv3)}
& \cellcolor{impStrong}0.9182 & \cellcolor{impStrong}0.0854
& \cellcolor{degMild}0.9293   & \cellcolor{degMild}0.0777
& \cellcolor{impStrong}0.8050 & \cellcolor{impStrong}0.1652
& \cellcolor{impStrong}0.7309 & \cellcolor{impStrong}0.2769
& \cellcolor{degStrong}0.6415 & \cellcolor{degStrong}0.3093 \\
{\nickname (Wan 2.1 14B, STream3R)}
& \cellcolor{impMild}0.8867 & \cellcolor{impMild}0.1164
& \cellcolor{impMild}0.9390  & \cellcolor{impMild}0.0715
& \cellcolor{impStrong}0.7855 & \cellcolor{impStrong}0.1762
& \cellcolor{impMild}0.5300   & \cellcolor{impMild}0.4391
& \cellcolor{impMild}0.7183   & \cellcolor{impMild}0.2379 \\
\bottomrule
\end{tabular*}
\end{table*}

\Cref{tab:ablation_progressive} ablates key design choices at a training and evaluation resolution of $256 \times 256$.

\paragraph{Importance of residual 3D tracking representation.}
Replacing \rtv{} with absolute 3D coordinate prediction (\textit{w/o Residual Pred}) causes a substantial drop across all benchmarks (\eg, APD on DR All $0.525 \to 0.296$).
Predicting absolute coordinates forces the model to memorize static scene geometry; the residual formulation factors out the static component and lets the VDM focus on dynamic deformation (Appendix \cref{fig:ablation_residual}).

\paragraph{Normalization and conditioning.}
Replacing log-space normalization with min-max (\textit{w/o Log Norm}) degrades performance, particularly on dynamic points (\eg, ADT Dyn $0.937 \to 0.920$): min-max compresses small displacements into a narrow range, whereas log-space expands them and maps zero displacement to zero.
Removing the point-map conditioning (\textit{w/o XYZ Cond}) degrades most on PStudio ($0.728 \to 0.640$) and DR Dyn ($0.701 \to 0.591$), supporting an explicit 3D scaffold.

\paragraph{Robustness to scaffold quality.}
DAv3~\cite{lin2026depth} yields stronger all-point tracking, while STream3R~\cite{stream3r2025} performs better on dynamic points, likely because DAv3 provides more accurate per-frame depth, benefiting static regions, while STream3R's causal architecture better preserves temporal consistency for moving objects.
The trade-off is largest on DR (dynamic-point APD $0.642 \to 0.718$ with STream3R, our default scaffold).

\paragraph{Scaling of video diffusion models.}
A key advantage of our framework is that it directly inherits advances in video generation. We compare 1.3B, 5B and 14B Wan backbones with identical hyperparameters. Larger models generally improve tracking, suggesting that stronger VDM priors translate into stronger tracking; with LoRA, even the 14B model trains on academic resources.

\subsection{Scaling the regression readout}%
\label{sec:regression-results}

\begin{figure}[t]
\centering
\includegraphics[width=\textwidth]{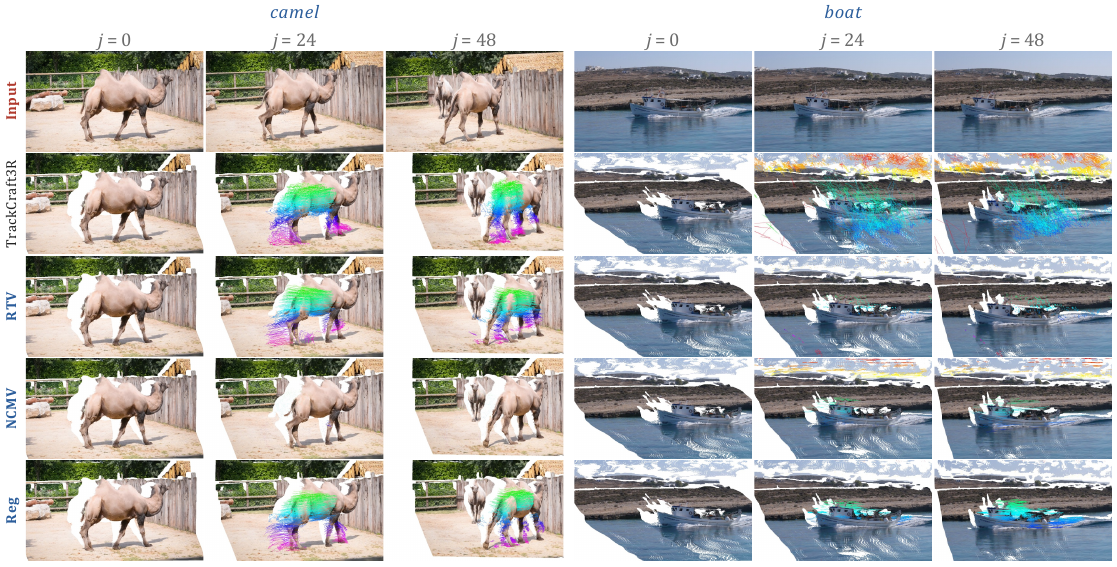}
\caption{\textbf{TrackCraft3R \vs the three \nickname{} readouts on DAVIS~\cite{davis_CVPR_2016}.}
Each panel shows the point cloud of frame $j$ and the last $12$ frames of the predicted 3D tracks, from one fixed view (the frame-$0$ input camera rotated by $4^\circ$). All methods receive the same VGGT-$\Omega$ point maps and draw the same $767$ queries, the most-moving points of a dense grid, colored by image row. \rtv{} and \ncmv{} come from the joint Wan2.1-I2V-14B model ($50$ steps, no input-latent noise); \ncmv{} tracks are read with CoTracker3, lifted with the same point maps, and frozen where \ncmv{} support is lost (on \emph{camel}, after the first frames). No ground truth or $\operatorname{Sim}(3)$ is used. More clips in \cref{fig:davis-cmp-more,fig:davis-cmp-more-b}.}%
\label{fig:davis-reg}
\end{figure}

The generative readout is limited by the video VAE (\cref{appendix:artifact-analysis}) and by dense supervision; we ask whether the regression readout of \cref{sec:regression} removes both limits once scaled, training it on eight dense and sparse sources that, as for V-DPM and 4RC, include PointOdyssey and Dynamic Replica (\cref{appendix:regression}).
We compare with TrackCraft3R~\cite{nam2026trackcraft3r}, which keeps the VAE output path, replaying its official checkpoint and \nickname-Reg under the scorer of \cref{tab:track_result} with the same VGGT-$\Omega$~\cite{wang26vggt-omega}, DAv3-L~\cite{lin2026depth} or STream3R point maps.

\Cref{tab:regression} shows that \nickname-Reg improves both metrics over TrackCraft3R in nine of twelve benchmark/geometry cells; EPE differs by less than $0.002$ in the other three.
Both systems, however, vary substantially with the source geometry, and no scaffold is best everywhere (\cref{appendix:oracle}); as the two models differ in backbone, raster, training data, loss and readout, the table compares complete systems.
\nickname-Reg ranks second behind V-DPM on PointOdyssey and Dynamic Replica (all points) and first on Dynamic Replica's dynamic points (\cref{tab:track_result}). The official V-DPM~\cite{sucar2025vdpm} and 4RC~\cite{luo264rc:} checkpoints, replayed under the same scorer with their own geometry, remain stronger on PointOdyssey and Dynamic Replica even against our best scaffold ($0.14$--$0.19$ against $0.21$--$0.23$ EPE, \cref{tab:regression}) and close on PStudio and ADT; both predict geometry jointly with motion, exactly where \nickname-Reg is weakest.
In \cref{fig:davis-reg}, its tracks stay on the boat, whereas TrackCraft3R's spread over the water and sky; over all $91$ DAVIS clips, the in-frame median second difference of the projected tracks is $1.8$ \vs $6.5$\,px, smaller for \nickname-Reg on $90$.

\paragraph{Geometry or motion?}
Substituting either factor of the output, the supplied query point or the predicted relative motion (\cref{eq:regression}), with ground truth \emph{after} prediction (\cref{tab:oracle-full}; a diagnostic only) shows that query geometry dominates the error on PointOdyssey and Dynamic Replica: EPE drops by an order of magnitude ($0.413 \to 0.053$ and $0.226 \to 0.027$ with VGGT-$\Omega$), versus at most $0.01$ for ground-truth motion. As a residual on the input point map, the readout is limited by that geometry, which suggests training geometry and motion jointly.

\section{Conclusions and Future Work}%
\label{sec:conclusions}

We presented \nickname, an early systematic study of how to represent dense 3D point tracking inside a pretrained video diffusion model.
We compared two natural representations within a single VDM\@: a Residual 3D Tracking Video (\rtv) from which metric tracks are read directly, and a Normalized Coordinate Map Video (\ncmv) that gives each 3D point a canonical color consistent over time.
Both are produced by LoRA adaptation of a frozen Wan backbone conditioned on point maps from an off-the-shelf 3D reconstructor.
Trained with a single dynamic 3D tracking dataset, \nickname is competitive across four zero-shot benchmarks and outperforms St4RTrack and TraceAnything by a clear margin on the held-out dense Kubric split.
Our central finding is that VDM priors transfer to dense 3D tracking when the right representation is chosen, and \rtv{} is the cleaner deployment path.
Without the output VAE, residual regression admits sparse supervision and, once scaled, outperforms TrackCraft3R on most benchmark/geometry pairs, showing the potential of this direction.
We view \nickname{} as an initial step toward repurposing VDM priors for 4D reconstruction.

\heading{Limitations and future work}
(1)~\rtv{} needs a target at every pixel, which restricts training to a single dense dynamic 3D tracking dataset; (2)~\ncmv{} yields less accurate tracks; (3)~\nickname-Reg is limited by its input geometry.
A natural next step is a unified model that predicts geometry and motion jointly, trained on large-scale 3D/4D data; see \cref{appendix:limitations} for further directions.

\begin{ack}
This research was supported by the Sony Research Award Programme 2024 and ERC CoG UNION (101001212). We thank Isambard-AI for support, and Yihang Luo and Gabrijel Boduljak for their discussion. The authors declare no competing interests.
\end{ack}

\FloatBarrier

{
\small
\setlength{\bibsep}{2pt plus 0.3ex}
\bibliographystyle{unsrt}

\bibliography{vedaldi_general,vedaldi_specific,yslan_general}
}

\appendix
\raggedbottom
\makeatletter
\setlength{\@fptop}{0pt}
\setlength{\@fpsep}{12pt}
\makeatother
\crefalias{section}{appendix}
\crefalias{subsection}{appendix}

\clearpage
\section{Limitations and future work}%
\label{appendix:limitations}

Our study exposes a trade-off between learnability and accuracy: (1)~\ncmv{} is expected to be easier for the VDM to learn but yields less accurate tracks; (2)~decoding \rtv{} through the video VAE introduces artifacts near object boundaries and needs a target at every pixel, which restricts training to a single dense dynamic 3D tracking dataset; (3)~a regression head that infers motion from VDM features, akin to inverse dynamics, admits sparse supervision and benefits from additional training data (\cref{sec:regression-results}).
We look forward to a unified model that predicts geometry and motion directly, trained on large-scale 3D/4D data, which requires more compute.

\nickname{} is trained on 49-frame clips (\nickname-Reg on 11-frame windows) and currently assumes the first frame as the query frame; extending it to longer videos and arbitrary query frames remains future work. The generative readouts do not predict per-point visibility. Faster sampling, \eg{} by consistency-style distillation, is another direction. Forecasting 3D tracks beyond the observed frames has been explored for scene flow~\cite{bharadhwaj26motionforesight} and point trajectories~\cite{zhang26molmomotion}; conversely, Track4Gen~\cite{jeong2025track4gen} shows that learning to track points improves video generation.

The generative readouts remain noisy at boundaries: the scaffold dominates global error, while the video VAE adds boundary artifacts (\cref{appendix:artifact-analysis}). Transformer-based VAEs may reduce these artifacts but require retraining~\cite{szymanowicz25bolt3d:}. Camera-following scenes can yield near-static RTV (\cref{fig:supp-failurecase}); the normalization audit preserves motion targets but does not establish the cause (\cref{appendix:camera-following-analysis}). Broader synthetic data~\cite{Black_CVPR_2023} or real-video pseudo-labels~\cite{uni4d_25_cvpr} could test the training-coverage hypothesis. We do not evaluate accuracy in occluded regions and make no claim of superiority there. \nickname-Reg predicts visibility, but we do not evaluate its accuracy. Direct feature readout lowers overall error and admits sparse supervision (\cref{appendix:direct-readout,sec:regression-results}), yet query geometry dominates the error of \nickname-Reg on PointOdyssey and Dynamic Replica (\cref{appendix:oracle}).

\section{Changes from the submitted version}
\label{appendix:changes}
\begin{figure}[h]
\centering
\includegraphics[width=\textwidth]{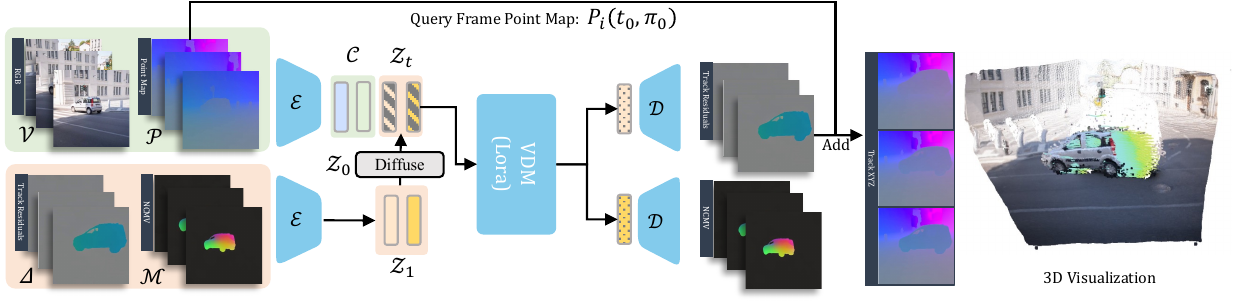}
\caption{\textbf{Submitted overview} (reproduced from the submitted version). A monocular video $\mathcal{V}$ and point maps $\mathcal{P}$ from an off-the-shelf reconstructor are encoded by the video VAE into the condition $\mathcal{C}$; the VDM jointly generates the \rtv{} $\bm{\Delta}$ and the \ncmv{} $\mathcal{M}$, and the decoded residuals are added to the query-frame point map. \Cref{fig:method_overview} shows the current overview.}
\label{fig:sub_overview}
\end{figure}

The task, title, and central question are unchanged: \nickname represents dense 3D tracking inside a pretrained video diffusion model, and compares a residual readout (\rtv{}) with a coordinate-map readout (\ncmv{}).
The generative model, its checkpoint, and every number in the submitted tables (main results, ablations, sampling steps) are unchanged.
This version adds a third readout that regresses the same residual without the video VAE, adds TrackCraft3R as a baseline, and expands the analysis.
\Cref{tab:changes} summarizes the differences and \cref{fig:sub_overview} reproduces the submitted overview.

\begin{table}[h]
\centering
\captionsetup{font=footnotesize}
\caption{\textbf{Submitted version versus this version.} The submitted column summarizes the described design; the training run that produced the submitted numbers is distinguished below.}
\label{tab:changes}
\footnotesize
\setlength{\tabcolsep}{4pt}
\begin{tabular}{@{}>{\raggedright\arraybackslash}p{0.17\linewidth}>{\raggedright\arraybackslash}p{0.385\linewidth}>{\raggedright\arraybackslash}p{0.385\linewidth}@{}}
\toprule
 & Submitted version & This version \\
\midrule
Readouts & \rtv{} and \ncmv{}, generated jointly by a LoRA-adapted Wan2.1-I2V-14B & Unchanged; adds \nickname-Reg, a regression readout of the \rtv{} residual without the output VAE (\cref{sec:regression}) \\
Generative model & Wan2.1-I2V-14B, LoRA rank $64$, point-map scaffold, Kubric and TartanAir & Unchanged checkpoint \\
Regression model & -- & Wan2.2-TI2V-5B with a motion expert and a DPT head, eight training sources (\cref{appendix:regression}) \\
Main table & Four sparse benchmarks and dense Kubric & Four sparse benchmarks; adds \nickname-Reg and TrackCraft3R; dense Kubric in \cref{tab:kubric} \\
Analyses & Ablations, sampling steps & Adds runtime, role of \ncmv{}, VAE-free readout, artifact sources, camera-following audit, geometry/motion oracle (\cref{appendix:additional-analysis}) \\
Figures & Overview of the generative model (\cref{fig:sub_overview}) & Overview redrawn with the regression readout (\cref{fig:method_overview}); DAVIS comparison with TrackCraft3R (\cref{fig:davis-reg,fig:davis-cmp-more}) \\
\bottomrule
\end{tabular}
\end{table}

\paragraph{Training details.}
The submitted text stated $10{,}000$ iterations on $4{\times}$A6000 GPUs with batch size $16$ and learning rate $2{\times}10^{-5}$.
The joint checkpoint behind the submitted and current numbers was continued from an RTV-only adapter for $5{,}000$ joint-training iterations on $32$ GH200 GPUs for about one day, with learning rate $4{\times}10^{-5}$; \cref{appendix:arch} now states this configuration.

\paragraph{Evaluation.}
Dense Kubric moved from the main table to \cref{tab:kubric}, since it is not part of the official WorldTrack evaluation.
TrackCraft3R~\cite{nam2026trackcraft3r} and \nickname-Reg are scored with the scorer of the main table; V-DPM and 4RC are replayed under the same scorer as references (\cref{tab:regression}).
Several of the analyses in \cref{appendix:additional-analysis} were first reported in the author response.

\paragraph{Presentation.}
The abstract, introduction, related work, and conclusion each gain sentences on the regression readout and on concurrent work; \cref{sec:regression,sec:regression-results} are new.
To keep the main text within ten pages, the residual-prediction ablation, additional DAVIS comparison, ablation table, and implementation details appear in the appendix. The main text retains both representation figures.

\section{Dataset details}%
\label{appendix:data}

The generative readouts require video data with \textbf{dense 3D tracking annotations} for training.
Existing 3D tracking datasets such as PointOdyssey~\cite{zheng2023point} and Dynamic Replica~\cite{karaev2023dynamicstereo} are not suitable: they provide only sparse annotations.
DELTA~\cite{ngo2024delta} provides a pre-processed Kubric set, but lacks flexibility in selecting the query frame (only predefined frames are queryable) and does not include the \ncmv{} representation; its sequences contain only $21$ frames, whereas our generative model uses $49$.

We re-render Kubric with the official Kubric-F~\cite{greff2021kubric} pipeline, redesigning the camera trajectories to introduce larger viewpoint variation.
The training set is $5{,}600$ sequences of $49$ frames each, with RGB, depth, camera, and dense 3D tracking annotations.
\ncmv{} targets are obtained by projecting query-frame 3D tracks onto the target camera trajectory.
We reserve a held-out evaluation split disjoint from training.
\ncmv{} videos are computed from the rendered tracks.
We additionally include TartanAir~\cite{tartanair2020iros} for static-scene diversity; for these sequences, the tracking ground truth is set to zero motion and \ncmv{} supervision is masked out.
We will release rendering scripts and the rendered training/evaluation splits.

\section{Architecture and LoRA details}%
\label{appendix:arch}

The generative readouts introduce no new output heads.
Both \rtv{} and \ncmv{} are 3-channel videos and are encoded by the original Wan VAE without modification --- the latent channel count is unchanged from the released VDM\@.
Multi-modality is handled at the token-sequence level: the clean target $\mathcal{Z}_1 = [x_{\tilde{\bm{\Delta}}};\, x_M]$ and the conditioning $\mathcal{C} = [x_\mathcal{V};\, x_\mathcal{P}]$ are each width-concatenated, then concatenated again to form the $4w$-wide DiT input (\cref{sec:model}), following~\cite{chen20254dnex,huang2026gen3r3dscenegeneration}.

LoRA~\cite{hu2022lora} is applied with rank $64$ to the attention and MLPs of the DiT\@;
the VAE encoder/decoder, the patch embedder, and the umT5~\cite{chung2023unimax} text encoder are kept frozen.
We add four learnable domain-specific tokens (one per modality: RGB, point map, \rtv{}, \ncmv{}) to differentiate input streams, following~\cite{chen20254dnex}.
Trainable parameters account for well under $1$\% of the DiT, allowing the 14B Wan backbone to be fine-tuned with plain data parallelism on $32$ GH200 GPUs.
We use a per-GPU batch of $2$ and learning rate $4{\times}10^{-5}$, with all other hyperparameters identical to the official Wan release; inputs are resized so the short side is up to $256$ pixels, with $49$ frames per clip.

The Wan implementation is adapted from DiffSynth-Studio with the video data-loading pipeline from STream3R~\cite{stream3r2025};
the multi-view augmentations originally designed for STream3R are removed.
For all loaded videos we follow Geo4D~\cite{geo4d}: we normalize the camera trajectory to the first-frame coordinate system and scale the point map by \texttt{common\_scale} (the mean absolute scale of $\mathcal{P}$).
Tracking offsets use the same geometric scale; the signed-log transform is inverted before adding predicted offsets to the query-frame point map. Undoing the geometric scaling then recovers metric trajectories.

\section{Comparison protocol details}%
\label{appendix:protocol}

\Cref{tab:track_result} compares baselines under their native protocols, each using its own depth/camera estimator.
Following D4RT~\cite{zhang2025d4rt}, each method runs at its native resolution: baselines resize internally, while the generative \nickname{} uses a $256$-pixel short side (\nickname-Reg: \cref{appendix:regression}).
We summarize the conditioning interfaces below for transparency:

\begin{itemize}
\item St4RTrack~\cite{st4rtrack2025}: jointly predicts tracking and depth from RGB\@; no external scaffold.
\item TraceAnything~\cite{liu2025traceanythingrepresentingvideo}: end-to-end from RGB\@.
\item Any4D~\cite{karhade2025any4d}: built on a VGGT-style 3D backbone trained on broad data.
\item V-DPM~\cite{sucar2025vdpm}: built on a VGGT-style 3D backbone trained on broad data and fine-tuned on multiple tracking benchmarks.
\item CoTracker3$^*$~\cite{cotracker3}: 2D tracker, lifted to 3D via VGGT~\cite{wang25vggt}.
\item SpatialTrackerV2~\cite{xiao2025spatialtracker}: joint depth + tracking head.
\item \nickname{} (ours): conditioned on point maps from STream3R~\cite{stream3r2025} or DAv3~\cite{lin2026depth}.
\end{itemize}


\paragraph{Note on SpatialTrackerV2.}
SpatialTrackerV2 also relies on scene-geometry conditioning;
ours uses a stronger 3D backbone (STream3R / DAv3) than its default depth pipeline.
Replacing SpatialTrackerV2's depth with our scaffold requires retraining its fusion head, which we do not pursue.


\section{Sampling steps}%
\label{appendix:sampling_steps}

VDMs typically require many denoising steps to generate high-quality videos.
Although~\nickname{} is built on a VDM backbone, it remains effective with substantially fewer sampling steps.
\Cref{tab:sampling_steps} shows that reducing the number of steps preserves, and in some cases improves, tracking performance while lowering inference time, consistent with observations in Geo4D~\cite{geo4d}.
Understanding and exploiting this sampling robustness for faster inference is an interesting direction for future work.
Unless otherwise stated, generative-model experiments use 50 sampling steps; \nickname-Reg uses no denoising.
The inference time is evaluated on a single NVIDIA A6000 GPU\@; peak GPU memory is $31.9$\,GB and is essentially independent of the sampling step count.

\begin{table}[t]
\centering
\captionsetup{font=footnotesize}
\caption{\textbf{Sampling-step ablation.}
We ablate inference-time sampling steps using a Wan2.2-5B \nickname{} model trained at $256{\times}256$ on $512$ Kubric scenes without point-map conditioning.
Performance remains largely stable when reducing the number of sampling steps from 50 to 10, with 5 steps still competitive while further reducing runtime.
}%
\label{tab:sampling_steps}
\renewcommand{\arraystretch}{1.05}
\setlength{\tabcolsep}{2pt}
\fontsize{8}{10}\selectfont
\begin{tabular*}{\linewidth}{@{\extracolsep{\fill}}c ccccc c@{}}
\toprule
\multirow{2}{*}{\textbf{Steps}}
& \multicolumn{5}{c}{\textbf{EPE}$\downarrow$}
& \multirow{2}{*}{\textbf{Time / Seq.\ $\downarrow$}} \\
\cmidrule(lr){2-6}
& Kubric~\cite{greff2021kubric}
& PO~\cite{zheng2023point}
& ADT~\cite{pan2023ariadigitaltwinnew}
& PStudio~\cite{pstudio}
& DR~\cite{karaev2023dynamicstereo}
& \\
\midrule
50
& 1.0463 & 0.3953 & 0.1149 & 0.2646 & 0.2549
& 14.0 $\pm$ 1.3\,s \\
10
& \textbf{1.0394} & 0.3821 & 0.1066 & \textbf{0.2642} & \textbf{0.2548}
& 9.6 $\pm$ 1.1\,s \\
5
& 1.0862 & \textbf{0.3773} & \textbf{0.1039} & 0.2668 & 0.2610
& \textbf{8.6} $\pm$ \textbf{1.1}\,s \\
\bottomrule
\end{tabular*}
\end{table}

\section{Additional experiments and analyses}%
\label{appendix:additional-analysis}

This section reports additional experiments and diagnostics motivated by questions about task-specific data efficiency, the role of \ncmv{}, computational cost, and the source of geometric artifacts. Unless noted, metrics use the main paper's closed-form $\operatorname{Sim}(3)$ alignment.

\subsection{Task-specific post-training data}%
\label{appendix:data-efficiency}

Both \nickname{} and V-DPM~\cite{sucar2025vdpm} adapt pretrained visual backbones rather than training them from scratch: \nickname{} builds on Wan, while V-DPM builds on VGGT. We therefore distinguish generic foundation-model pretraining from task-specific post-training. The dynamic component of generative \nickname{} uses $5{,}600\times49\approx274$K densely annotated Kubric frames. V-DPM instead uses four dynamic sources (Kubric-F, Kubric-G, PointOdyssey, and Waymo), alongside two static sources (ScanNet++ and BlendedMVS). Its reported per-epoch sampling counts do not establish a directly comparable total of distinct annotated frames, so we do not infer a frame-count ratio. This comparison does not claim lower total pretraining cost, since the generic pretraining data and compute of Wan and VGGT are not directly comparable.

The distinction also affects benchmark interpretation. PointOdyssey is included in the training data of V-DPM, SpatialTrackerV2, St4RTrack, and Any4D, but is fully held out for the submitted \nickname{} model. The PointOdyssey results of the generative readouts in \cref{tab:track_result} therefore measure zero-shot transfer, whereas this benchmark is in-domain for these baselines and \nickname-Reg.

\subsection{Dense Kubric evaluation}%
\label{appendix:kubric}

\Cref{tab:kubric} reports the held-out dense Kubric split. It matches the training distribution of \nickname-RTV and is reported separately from the WorldTrack benchmarks in \cref{tab:track_result}.

\begin{table}[!htbp]
\centering
\captionsetup{font=footnotesize}
\caption{\textbf{Dense 3D tracking on the held-out Kubric split.}
APD$\uparrow$ / EPE$\downarrow$ after $\operatorname{Sim}(3)$ alignment on $50$ held-out sequences of $49$ frames, dense annotations; ``All'' and ``Dyn'' as in \cref{tab:track_result}.
Kubric matches the training distribution of \nickname-RTV and is not part of the benchmark suite of \cref{tab:track_result}.}%
\label{tab:kubric}
\renewcommand{\arraystretch}{1.05}
\setlength{\tabcolsep}{2pt}
\fontsize{8}{10}\selectfont
\begin{tabular*}{\linewidth}{@{\extracolsep{\fill}}l cc cc@{}}
\toprule
\multirow{2}{*}{\textbf{Method}} & \multicolumn{2}{c}{All} & \multicolumn{2}{c}{Dyn} \\
\cmidrule(lr){2-3} \cmidrule(lr){4-5}
& APD$\uparrow$ & EPE$\downarrow$ & APD$\uparrow$ & EPE$\downarrow$ \\
\midrule
St4RTrack~\cite{st4rtrack2025} & 0.0468 & 5.3331 & 0.0546 & 2.8485 \\
TraceAnything~\cite{liu2025traceanythingrepresentingvideo} & \underline{0.2879} & \underline{1.5501} & 0.1379 & 1.9263 \\
Any4D~\cite{karhade2025any4d} & 0.2242 & 1.9223 & 0.1370 & 1.8107 \\
V-DPM~\cite{sucar2025vdpm} & 0.1729 & 4.7054 & \textbf{0.4822} & \textbf{0.6845} \\
\rowcolor{gray!10}[0pt][0pt]\oursrowshade \nickname-RTV & \textbf{0.3359} & \textbf{1.0790} & \underline{0.1890} & \underline{1.4042} \\
\bottomrule
\end{tabular*}
\end{table}

\subsection{Inference time and memory}%
\label{appendix:efficiency-comparison}

For the baseline timings, we measure inference cost on ten fixed $49$-frame clips using each method's official preprocessing and native inference resolution. After warm-up, we record the mean and standard deviation of per-sequence runtime and the peak reserved GPU memory. \Cref{tab:runtime-comparison} uses the $5$-step \nickname{} configuration, which retains similar EPE to the default $50$-step sampler (\cref{tab:sampling_steps}).

\begin{table}[!htbp]
\centering
\captionsetup{font=footnotesize}
\caption{\textbf{Inference time and peak memory for $49$-frame sequences on one GPU.}
Native inference resolutions are shown in parentheses. The \nickname{} row measures the $5$B, $5$-step configuration, which is not the configuration used for the accuracy results in \cref{tab:track_result}. V-DPM does not fit on the $48$\,GB A6000; its H100 measurement is included only as a hardware-unmatched reference and should not be read as a direct runtime comparison. $^\dagger$Measured on an A100 over $90$ warm DAVIS clips (five passes, BF16), excluding the production of the input point maps; also hardware-unmatched.}
\label{tab:runtime-comparison}
\renewcommand{\arraystretch}{1.05}
\setlength{\tabcolsep}{2pt}
\fontsize{8}{10}\selectfont
\begin{tabular*}{\linewidth}{@{\extracolsep{\fill}}lccc@{}}
\toprule
\textbf{Method} & \textbf{Hardware} & \textbf{Runtime (s/seq.) $\downarrow$} & \textbf{Peak memory (GB, reserved) $\downarrow$} \\
\midrule
St4RTrack ($512^2$) & A6000 48\,GB & $6.07 \pm 0.03$ & \textbf{9.56} \\
TraceAnything ($512^2$) & A6000 48\,GB & $6.76 \pm 0.05$ & 37.17 \\
CoTracker + VGGT ($476^2$) & A6000 48\,GB & $7.53 \pm 0.15$ & 23.04 \\
4RC ($504\times378$) & A6000 48\,GB & $8.76 \pm 0.09$ & 20.60 \\
Any4D ($518^2$) & A6000 48\,GB & $10.07 \pm 0.14$ & 43.31 \\
SpatialTrackerV2 ($518^2$) & A6000 48\,GB & $28.00 \pm 2.71$ & 26.29 \\
V-DPM ($518^2$) & A6000 48\,GB & OOM & -- \\
V-DPM ($518^2$) & H100 80\,GB & 139.39 & 53.49 \\
\textbf{\nickname{} (5B, 5 steps)} & A6000 48\,GB & $8.60 \pm 1.10$ & 31.9 \\
\textbf{\nickname-Reg} ($288{\times}512$)$^\dagger$ & A100 80\,GB & $6.51 \pm 0.51$ & 26.0 \\
\bottomrule
\end{tabular*}
\end{table}

On the shared A6000 hardware, \nickname{} is slower and more memory-intensive than the lightest feed-forward baselines, but has similar runtime to 4RC and Any4D in this suite. Its $31.9$\,GB peak reserved memory is lower than TraceAnything and Any4D, but higher than St4RTrack, CoTracker+VGGT, 4RC, and SpatialTrackerV2. V-DPM exceeds the A6000 memory budget.

\subsection{Role and supervision of \ncmv{}}%
\label{appendix:ncmv-analysis}

\paragraph{Foreground supervision.}
Kubric provides ground-truth instance masks. During training, we supervise \ncmv{} only on foreground objects; for the static TartanAir clips, \ncmv{} supervision is masked out. At inference, the model receives no external foreground mask. It predicts \ncmv{} directly and must implicitly identify the relevant foreground regions. This separation is often successful, but it can fail when a moving object is interpreted as static.

\paragraph{RTV-only versus joint training.}
We compare the Wan-14B RTV-only checkpoint with its joint RTV+NCMV continuation in \cref{tab:rtv-only-ablation}. The joint checkpoint improves EPE and APD on every benchmark, reducing mean all-point EPE by $2.34\%$ and mean dynamic EPE by $2.93\%$ (computed from unrounded means). Training history and sampling settings are not fully matched, so this is a checkpoint comparison rather than an isolated NCMV-loss ablation.

\begin{table}[!htbp]
\centering
\captionsetup{font=footnotesize}
\caption{\textbf{RTV-only and joint RTV+NCMV checkpoints.}
Per-benchmark rows report dynamic EPE $\downarrow$ / APD $\uparrow$ on $49$-frame sequences; the final row reports all-point means. Both are Wan-14B LoRA checkpoints with different training histories, and they are not sampled identically. The joint checkpoint is the model of \cref{tab:track_result}: here it is sampled with $1$ denoising step and CFG $1$ at a $256$-pixel short side, whereas \cref{tab:track_result} reports its submitted evaluation with $50$ denoising steps and random input-latent noise (\cref{sec:exp-settings}). The two tables therefore give different values for the same checkpoint (\eg PStudio dynamic EPE $0.1511$ here \vs $0.1596$ in \cref{tab:track_result}). Bold denotes the better result.}
\label{tab:rtv-only-ablation}
\renewcommand{\arraystretch}{1.05}
\setlength{\tabcolsep}{2pt}
\fontsize{8}{10}\selectfont
\begin{tabular*}{\linewidth}{@{\extracolsep{\fill}}lcc@{}}
\toprule
\textbf{Benchmark} & \textbf{RTV-only} & \textbf{Joint RTV+NCMV} \\
\midrule
PointOdyssey & 0.2891 / 0.6908 & \textbf{0.2856 / 0.6931} \\
ADT & 0.0715 / 0.9390 & \textbf{0.0666 / 0.9466} \\
Panoptic Studio & 0.1610 / 0.8093 & \textbf{0.1511 / 0.8240} \\
Dynamic Replica & 0.1886 / 0.7813 & \textbf{0.1861 / 0.7850} \\
\midrule
Mean (dynamic) & 0.1776 / 0.8051 & \textbf{0.1724 / 0.8122} \\
Mean (all-point) & 0.1848 / 0.8134 & \textbf{0.1804 / 0.8195} \\
\bottomrule
\end{tabular*}
\end{table}

The joint recipe improves \rtv{} despite generating an additional representation. The table reports one checkpoint pair, not a multi-seed study, and does not isolate the auxiliary-task contribution. We retain \rtv{} as the primary practical output and regard \ncmv{} as an auxiliary representation analysis and training objective. Because \ncmv{} assigns persistent canonical coordinates rather than anchoring every output to a single query frame, it can in principle represent points that first become visible later. Our targets, however, are built from query-frame tracks, so exploiting this would require canonical-coordinate supervision for such points; it has not been validated empirically here.

\paragraph{A dedicated $512\times512$ NCMV model.}
For visualization we also train a separate \ncmv{}-only model. It adapts Wan2.1-T2V-14B with rank-$64$ LoRA, is trained only on Kubric \ncmv{} targets ($5{,}636$ clips, $512\times512$) with a two-phase curriculum, $4{,}000$ steps on $29$-frame windows and then $4{,}000$ steps on full $49$-frame clips (effective batch $16$, about $270$ H200 GPU-hours in total), and samples \ncmv{} in $50$ denoising steps with source-video classifier-free guidance $3$. Unlike the joint model, it is conditioned on the RGB video alone, without point maps. We use it only for qualitative results on the project page; every \ncmv{} number in this paper comes from the joint model.

\subsection{VAE-free direct feature readout}%
\label{appendix:direct-readout}

The original formulation encodes the dense tracking target as a three-channel video and decodes it through the RGB-pretrained video VAE. This requires a tracking value at every pixel and therefore prevents direct use of datasets that provide only sparse trajectories. Motivated by shallow readouts for measuring the 3D information in video-foundation-model features~\cite{huang26how-much-3d}, we attach a shallow ViT/DPT-style regression head to intermediate Wan2.2-5B features and jointly train the head with LoRA adaptation of the backbone. The regression loss is applied only at annotated locations, making sparse tracking datasets compatible with the framework.

\Cref{tab:direct-readout} separates the effects of the output path, auxiliary reconstruction supervision, and additional tracking data. Replacing VAE decoding with direct readout under the same Kubric tracking data lowers mean dynamic EPE from $0.232$ to $0.225$. Adding auxiliary depth and camera regression lowers it to $0.212$, and adding PointOdyssey tracks lowers it further to $0.186$. The complete recipe improves mean dynamic EPE by approximately $20\%$ and raises mean APD from $0.740$ to $0.793$.

\begin{table}[!htbp]
\centering
\captionsetup{font=footnotesize}
\caption{\textbf{VAE-free direct readout with a Wan2.2-5B backbone.}
Dynamic EPE $\downarrow$ / APD $\uparrow$; the mean averages the four benchmarks. The first two rows compare the output paths under the same Kubric tracking data. The final row additionally uses PointOdyssey for training, so PointOdyssey is in-domain for that variant.}
\label{tab:direct-readout}
\renewcommand{\arraystretch}{1.05}
\setlength{\tabcolsep}{2pt}
\fontsize{8}{10}\selectfont
\begin{tabular*}{\linewidth}{@{\extracolsep{\fill}}lccccc@{}}
\toprule
\textbf{Model} & \textbf{PointOdyssey} & \textbf{ADT} & \textbf{PStudio} & \textbf{DynamicReplica} & \textbf{Mean} \\
\midrule
Submitted 5B (VAE decode) & 0.379 / 0.594 & \textbf{0.072 / 0.937} & 0.223 / 0.728 & 0.254 / 0.701 & 0.232 / 0.740 \\
Direct readout, Kubric tracking & 0.321 / 0.655 & 0.117 / 0.871 & 0.251 / 0.695 & 0.210 / 0.753 & 0.225 / 0.743 \\
$+$ depth and camera supervision & 0.317 / 0.657 & 0.107 / 0.883 & 0.216 / 0.734 & 0.206 / 0.758 & 0.212 / 0.758 \\
$+$ PointOdyssey tracking & \textbf{0.290 / 0.689} & 0.090 / 0.913 & \textbf{0.176 / 0.786} & \textbf{0.187 / 0.785} & \textbf{0.186 / 0.793} \\
\bottomrule
\end{tabular*}
\end{table}

The $0.232\rightarrow0.225$ comparison is the data-matched comparison of the output paths. The full $0.232\rightarrow0.186$ gain is cumulative and should not be attributed to direct readout alone. In particular, adding PointOdyssey to the model with auxiliary reconstruction supervision provides a further $12.3\%$ EPE reduction. These experiments are currently limited to the $5$B backbone.

\subsection{Regression readout: architecture, training and protocol}%
\label{appendix:regression}

\begin{figure}[t]
\centering
\includegraphics[width=1.0\textwidth]{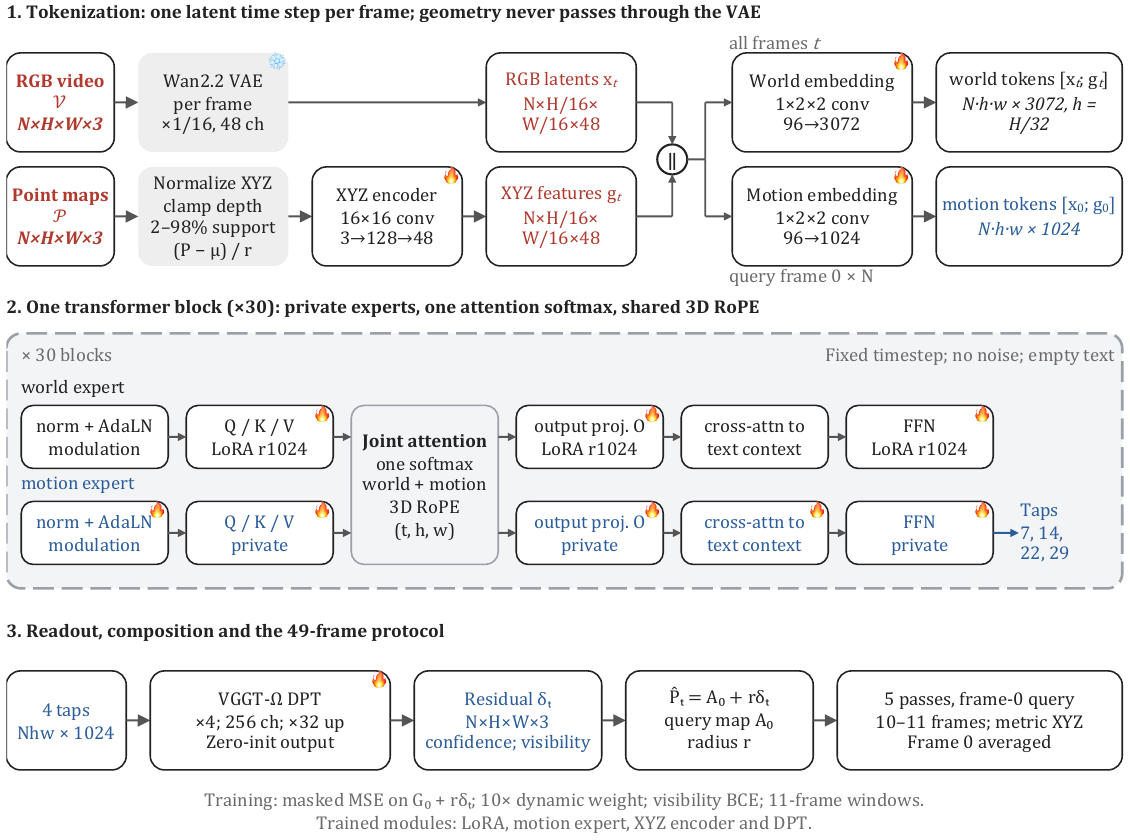}
\caption{\textbf{Tensor flow of the regression readout (\nickname-Reg).}
\textbf{Top:} frame-wise RGB latents and point-map features form world tokens; repeating the query-frame features forms motion tokens.
\textbf{Middle:} the LoRA-adapted Wan expert and private motion expert share attention and 3D rotary positions in each block.
\textbf{Bottom:} motion features from four blocks feed a DPT head for residuals and visibility. Residuals are added to the supplied query point map; five frame-0-anchored passes cover $49$ frames.
Flames: trained modules; VAE and Wan base weights: frozen.}%
\label{fig:supp-mot-arch}
\end{figure}

This section documents the scaled regression model of \cref{sec:regression} (\nickname-Reg), whose results are reported in \cref{sec:regression-results}.
It is a different model family from the generative \rtv{}/\ncmv{} models of \cref{tab:track_result} and from the $5$B direct-readout probes of \cref{appendix:direct-readout}; the three must not be read as one ablation ladder, and the STream3R rows in \cref{tab:track_result,tab:regression} report the same evaluation.

\paragraph{Architecture.}
\Cref{fig:supp-mot-arch} shows the tensor flow.
The RGB video is encoded frame by frame by the frozen Wan2.2 VAE ($\times 1/16$, $48$ channels), so that every frame keeps its own latent time step; the normalized point maps are encoded per frame by a $16{\times}16$ patch encoder ($3\to128\to48$ channels) and never pass through the VAE\@.
The two $48$-channel maps are concatenated and enter the Wan2.2-TI2V-5B DiT ($30$ blocks, width $3{,}072$) through its patch embedding, widened from $48$ to $96$ input channels with the geometry half initialized from the RGB weights; this is the \emph{world expert}, adapted with LoRA of rank $1{,}024$ on the Q/K/V/O and FFN projections.
The \emph{motion expert} is a private $30$-block transformer (width $1{,}024$, FFN width $4{,}096$) with its own patch embedding, norms, modulation, Q/K/V/O and FFN\@; its input is the query-frame pair $[x_{\mathcal{V}}; g_{\mathcal{P}}]_0$ repeated at every target time~\cite{nam2026trackcraft3r}.
In each block the queries, keys and values of both groups are projected into the same head space, carry the same 3D rotary position (time, height, width), and are concatenated under one attention softmax~\cite{liang2024mixture}; each group then applies its own output projection, cross-attention to the (empty) text context and FFN\@.
Motion tokens after blocks $7$, $14$, $22$ and $29$ feed the VGGT-$\Omega$ DPT head~\cite{ranftl21vision,wang26vggt-omega} ($256$ fusion features, $\times 32$ spatial upsampling, zero-initialized output projection so that training starts from the zero-motion predictor), which outputs the residual (identity activation), a confidence and a separate visibility logit.
The query-frame residual is supervised like any other frame and is not forced to zero at inference.
Inference uses the last training timestep as a fixed regression timestep, with no noise, classifier-free guidance or iterative denoising; the world blocks after the deepest tap cannot reach the readout and their LoRA tensors are frozen.
The checkpoint holds $1{,}475$ trained tensors ($588$ LoRA and patch-embedding, $822$ motion expert, $62$ DPT, $3$ patch encoder); the base Wan weights are loaded unchanged.

\paragraph{Normalization and composition.}
Each clip's point maps are expressed in the first camera, depth is clamped to $[10^{-6}, 80]$\,m, the dense support between the 2nd and 98th depth percentiles is centred on its mean $\mu$ and divided by its maximum radius $r$.
With $a_0 = (\pzero{0}{0}-\mu)/r$ the normalized query point and $\bm{\delta}_j$ the predicted residual, the metric output is $\hat{P}_0(t_j) = \mu + r(a_0 + \bm{\delta}_j) = \pzero{0}{0} + r\,\bm{\delta}_j$ (\cref{eq:regression}).
During training the loss composes the residual with the \emph{ground-truth} query point, $G_0 + r\,\bm{\delta}_j$, whereas deployment composes it with the \emph{estimated} point map; the model is therefore not trained to correct an imperfect anchor, which the oracle analysis of \cref{appendix:oracle} identifies as the dominant error source on PointOdyssey and Dynamic Replica.

\paragraph{Training.}
Windows of $11$ frames are sampled in sequential order with per-source maximum frame intervals between $4$ and $9$ and a reversal probability of $0.25$, at three canvases ($512{\times}288$, $384{\times}384$, $288{\times}512$).
The mixture contains our Kubric split and Kubric-F~\cite{greff2021kubric}, PointOdyssey~\cite{zheng2023point}, Dynamic Replica~\cite{karaev2023dynamicstereo}, Waymo~\cite{waymo}, DL3DV~\cite{ling2024dl3dv}, ScanNet~\cite{dai2017scannet} and MVS-Synth~\cite{DeepMVS}; the loss is applied only at annotated pixels (masked MSE on the composed points with a $10\times$ weight on dynamic pixels and a visibility BCE), so sparse sources are used as is.
The reported checkpoint is the end of a three-stage, weights-only warm-start lineage ($12{,}750 \to 500 \to 1{,}000$ steps, optimizer and RNG state reset at each boundary) and is not a model trained from scratch for $1{,}000$ steps; all three stages ran on two NVIDIA H200 GPUs, about seven days in total; the final stage uses two processes with a per-process batch of $32$ and gradient accumulation $2$ (effective batch $128$), AdamW with learning rate $5{\times}10^{-5}$ (minimum $10^{-5}$), betas $0.9/0.95$, visibility loss weight $1$, and bf16 checkpoints; RGB latent noise and the image/geometry augmentations are disabled in this stage.

\begin{table}[htbp]
\centering
\captionsetup{font=footnotesize}
\caption{\textbf{Regression readout \vs TrackCraft3R.}
All-point EPE$\downarrow$ / APD$\uparrow$ on $50$ sequences per benchmark ($49$ for ADT), using the protocol of \cref{tab:track_result}. TrackCraft3R and \nickname-Reg share point-map caches; bold marks the better result. V-DPM and 4RC use their own geometry. PointOdyssey and Dynamic Replica are training datasets for \nickname-Reg, V-DPM and 4RC.
$^\circ$Oracle diagnostic without retraining: the ground-truth query point replaces the estimated one after prediction, scored under the fixed $\operatorname{Sim}(3)$ of the row above (\cref{tab:oracle-full}), following TrackCraft3R's input-geometry ablation.
}%
\label{tab:regression}
\renewcommand{\arraystretch}{1.05}
\setlength{\tabcolsep}{2pt}
\fontsize{8}{10}\selectfont
\begin{tabular*}{\linewidth}{@{\extracolsep{\fill}}ll cccc@{}}
\toprule
\textbf{Geometry} & \textbf{Method} & \textbf{PStudio} & \textbf{PointOdyssey} & \textbf{Dynamic Replica} & \textbf{ADT} \\
\midrule
\multirow{2}{*}{own}
& V-DPM~\cite{sucar2025vdpm} & 0.0883 / 0.9224 & 0.1403 / 0.8642 & 0.1914 / 0.8184 & 0.1045 / 0.9024 \\
& 4RC~\cite{luo264rc:} & 0.1067 / 0.8936 & 0.1939 / 0.8347 & 0.1604 / 0.8664 & 0.1144 / 0.8894 \\
\midrule
& TrackCraft3R~\cite{nam2026trackcraft3r} & 0.1017 / 0.9031 & 0.4370 / 0.6524 & \textbf{0.2239} / \textbf{0.7783} & 0.0935 / \textbf{0.9245} \\
\rowcolor{gray!10}[0pt][0pt]\oursrowshade & \nickname-Reg (ours) & \textbf{0.0947} / \textbf{0.9155} & \textbf{0.4127} / \textbf{0.6668} & 0.2257 / 0.7777 & \textbf{0.0933} / 0.9230 \\
\multirow{-3}{*}{VGGT-$\Omega$} & \textcolor{gray}{\nickname-Reg + GT query geometry$^\circ$} & \textcolor{gray}{0.0900 / 0.9169} & \textcolor{gray}{0.0531 / 0.9450} & \textcolor{gray}{0.0275 / 0.9743} & \textcolor{gray}{0.0374 / 0.9582} \\
\midrule
& TrackCraft3R~\cite{nam2026trackcraft3r} & 0.1347 / 0.8491 & 0.5224 / 0.5064 & 0.2859 / 0.7134 & 0.4002 / 0.6187 \\
\rowcolor{gray!10}[0pt][0pt]\oursrowshade & \nickname-Reg (ours) & \textbf{0.1270} / \textbf{0.8625} & \textbf{0.3536} / \textbf{0.6616} & \textbf{0.2836} / \textbf{0.7182} & \textbf{0.1400} / \textbf{0.8498} \\
\multirow{-3}{*}{DAv3-L} & \textcolor{gray}{\nickname-Reg + GT query geometry$^\circ$} & \textcolor{gray}{0.1146 / 0.8761} & \textcolor{gray}{0.0970 / 0.8944} & \textcolor{gray}{0.0337 / 0.9666} & \textcolor{gray}{0.0555 / 0.9401} \\
\midrule
& TrackCraft3R~\cite{nam2026trackcraft3r} & 0.1400 / 0.8418 & 0.2227 / 0.7904 & 0.2289 / 0.7839 & \textbf{0.1163} / \textbf{0.8874} \\
\rowcolor{gray!10}[0pt][0pt]\oursrowshade \multirow{-2}{*}{STream3R} & \nickname-Reg (ours) & \textbf{0.1311} / \textbf{0.8572} & \textbf{0.2078} / \textbf{0.8050} & \textbf{0.2284} / \textbf{0.7871} & 0.1168 / 0.8854 \\
\bottomrule
\end{tabular*}
\end{table}

\begin{figure}[!tb]
\centering
\includegraphics[width=\textwidth]{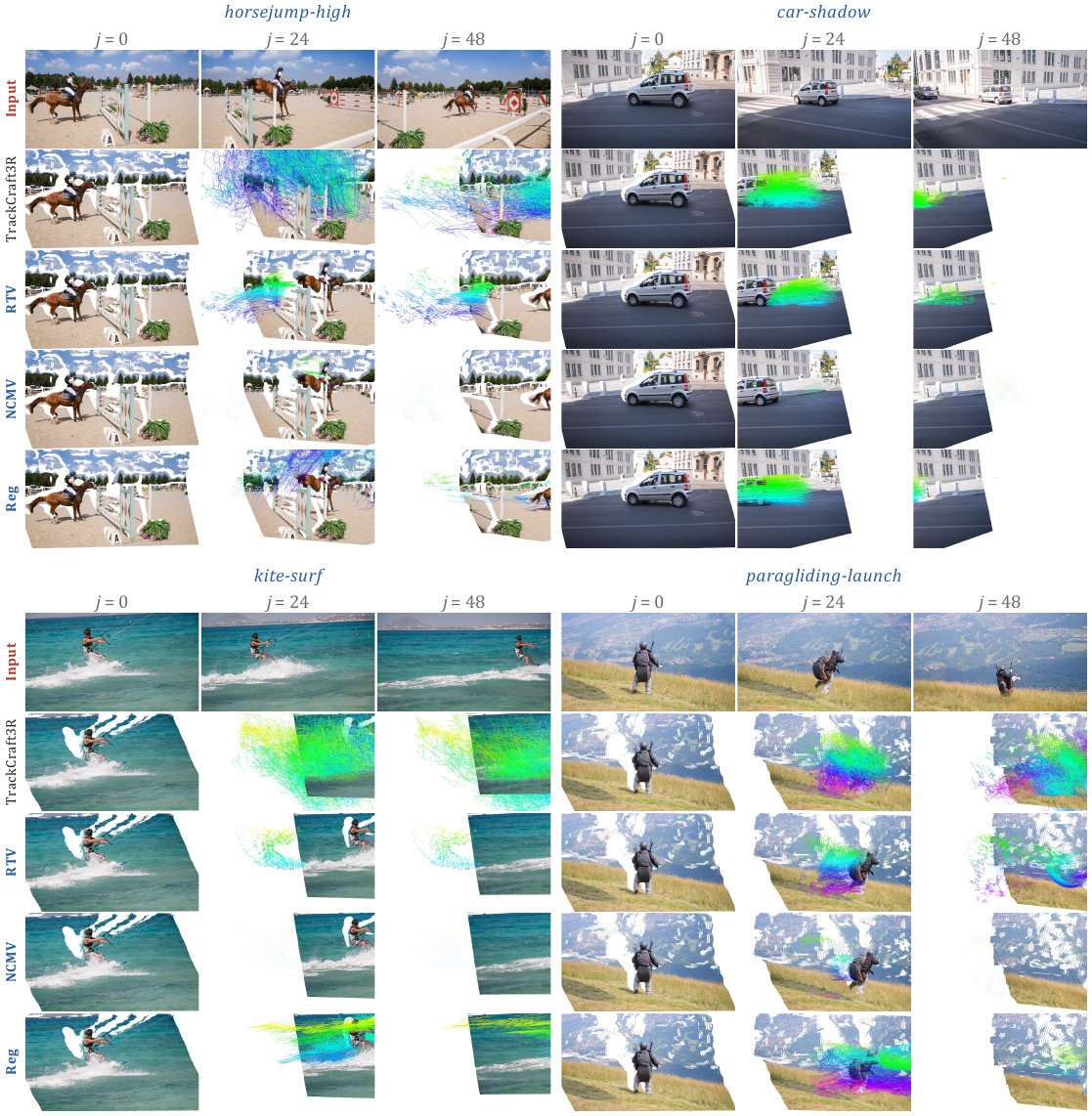}
\caption{\textbf{More DAVIS clips, TrackCraft3R \vs the three \nickname{} readouts.} Same setup as \cref{fig:davis-reg}. Because the view is fixed at frame $0$, subjects that the camera follows leave the view in later frames (\emph{car-shadow}, \emph{kite-surf}, \emph{parkour}).
Over all $91$ DAVIS clips, the median second difference of the projected tracks over consecutive frames, counting points inside the frame, is $6.5$\,px for TrackCraft3R and $1.8$\,px for \nickname-Reg ($672$-px-wide frames), and it is smaller for \nickname-Reg on $90$ clips. This measures temporal smoothness, not accuracy.}%
\label{fig:davis-cmp-more}
\end{figure}
\begin{figure}[!tb]
\centering
\includegraphics[width=\textwidth]{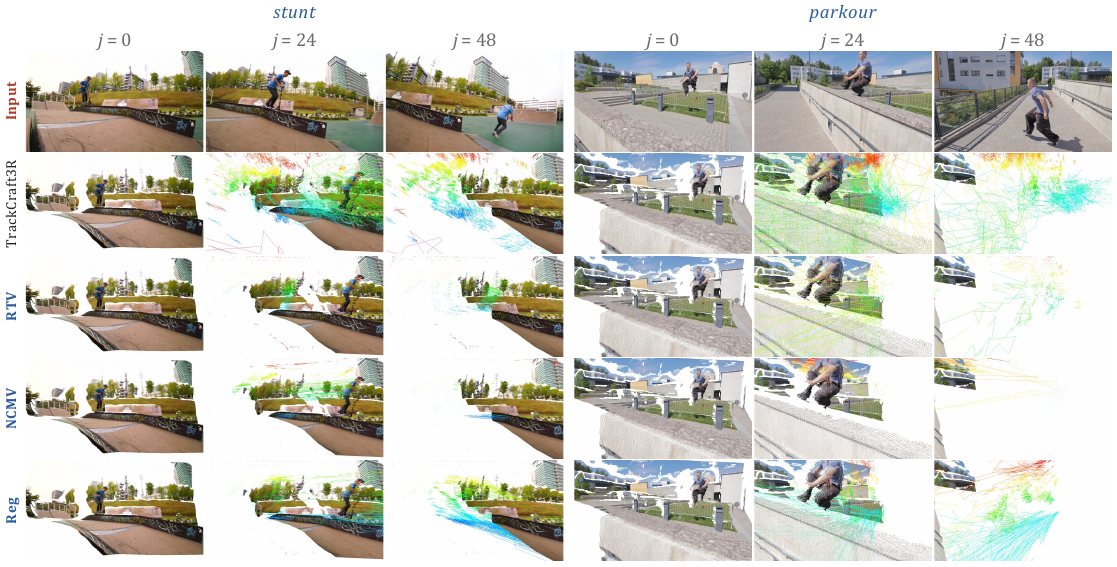}
\caption{\textbf{More DAVIS clips (continued).} Same setup as \cref{fig:davis-reg,fig:davis-cmp-more}.}%
\label{fig:davis-cmp-more-b}
\end{figure}

\paragraph{Evaluation protocol.}
\Cref{tab:regression,tab:oracle-full} use $50$ sequences per benchmark ($49$ for ADT, whose one sequence without a valid frame-0 query is excluded by the membership rule), the first $49$ frames, queries at ground-truth points visible in frame~0 mapped to the model raster by the width/height ratio and sampled at the nearest pixel, and one closed-form $\operatorname{Sim}(3)$ per sequence over all selected tracks and timestamps; APD uses thresholds $\{0.1, 0.3, 0.5, 1.0\}$\,m.
\nickname-Reg runs at $288{\times}512$ ($384{\times}384$ for ADT) and covers the $49$ frames with five deterministic frame-0-anchored passes of $10$ or $11$ frames (frame IDs $\{0,5,10,\dots\}$, $\{0,1,6,11,\dots\}$, $\{0,2,7,12,\dots\}$, $\{0,3,8,13,\dots\}$, $\{0,4,9,14,\dots\}$); frame~0 is averaged over the passes and every other frame is predicted once.
Each pass is normalized and mapped back to metric coordinates independently before merging; there is no temporal smoothing, per-pass alignment or interpolation.
The TrackCraft3R rows replay the official checkpoint and preprocessing at its native $480{\times}832$ raster with the same point-map caches and scorer; the comparison therefore matches data membership and scorer but not model input raster.
The V-DPM and 4RC rows replay the official checkpoints with their own geometry pipelines under the same membership and scorer ($49$ frames, query frame $0$), so they are references for the absolute level of the benchmark rather than a controlled comparison.
The geometry caches come from VGGT-$\Omega$~\cite{wang26vggt-omega}, DAv3-L (Depth Anything 3, Large)~\cite{lin2026depth} and STream3R~\cite{stream3r2025}.

\paragraph{Comparison with the generative models.}
\Cref{tab:reg-vs-generative} scores \nickname-Reg with STream3R geometry, the scaffold of \cref{tab:track_result}, against the two generative models of the paper under the same scorer: the Wan2.2-5B ablation baseline of \cref{tab:ablation_progressive} and the joint Wan2.1-I2V-14B model of \cref{tab:track_result} ($1$-step dynamic values of \cref{tab:rtv-only-ablation}; with the $50$-step values of \cref{tab:track_result} the ranking below is unchanged).
\nickname-Reg is better than both on PStudio, PointOdyssey and Dynamic Replica and is close to the 14B model on ADT, with a $5$B backbone and one forward pass per window.
The comparison is not controlled: the generative models run at a $256$-pixel short side and see only Kubric and TartanAir, whereas \nickname-Reg runs at $288{\times}512$ and trains on eight sources including PointOdyssey and Dynamic Replica, so PStudio and ADT are the only zero-shot cells for all three.

\begin{table}[!htbp]
\centering
\captionsetup{font=footnotesize}
\caption{\textbf{\nickname-Reg against the generative models with the same STream3R scaffold.}
Dynamic EPE$\downarrow$ / APD$\uparrow$ under the WorldTrack scorer ($50$ sequences, $49$ frames; ADT $49$); the last column reports all-point values of \nickname-Reg. Bold marks the best dynamic value. See the text for the differences in raster and training data.}%
\label{tab:reg-vs-generative}
\renewcommand{\arraystretch}{1.05}
\setlength{\tabcolsep}{2pt}
\fontsize{8}{10}\selectfont
\begin{tabular*}{\linewidth}{@{\extracolsep{\fill}}lcccc@{}}
\toprule
\textbf{Benchmark} & \textbf{Wan2.2-5B (\rtv)} & \textbf{Wan2.1-14B joint (\rtv)} & \textbf{\nickname-Reg} & \textbf{\nickname-Reg, all points} \\
\midrule
PStudio         & 0.2231 / 0.7281 & 0.1511 / 0.8240 & \textbf{0.1311} / \textbf{0.8572} & 0.1311 / 0.8572 \\
PointOdyssey    & 0.3792 / 0.5939 & 0.2856 / 0.6931 & \textbf{0.2691} / \textbf{0.7155} & 0.2078 / 0.8050 \\
Dynamic Replica & 0.2536 / 0.7009 & 0.1861 / 0.7850 & \textbf{0.1674} / \textbf{0.8141} & 0.2284 / 0.7871 \\
ADT             & 0.0722 / 0.9370 & \textbf{0.0666} / \textbf{0.9466} & 0.0675 / 0.9430 & 0.1168 / 0.8854 \\
\bottomrule
\end{tabular*}
\end{table}

\subsection{Geometry-versus-motion oracle}%
\label{appendix:oracle}

\begin{table}[!htbp]
\centering
\captionsetup{font=footnotesize}
\caption{\textbf{Full geometry/motion oracle for \nickname-Reg.}
EPE$\downarrow$ / APD$\uparrow$ on all $50$ sequences of each primary benchmark (ADT, $49$ sequences, is secondary).
\emph{Fixed} columns score the substituted output under the original all-track $\operatorname{Sim}(3)$ of the current prediction; \emph{refit} columns fit a new closed-form $\operatorname{Sim}(3)$ to each substituted output.
Closed-form $\operatorname{Sim}(3)$ minimizes squared distances rather than EPE or APD, so refitting can raise EPE; neither column gives a lower bound.}%
\label{tab:oracle-full}
\renewcommand{\arraystretch}{1.05}
\setlength{\tabcolsep}{2pt}
\fontsize{8}{10}\selectfont
\begin{tabular*}{\linewidth}{@{\extracolsep{\fill}}ll c cc cc@{}}
\toprule
& & & \multicolumn{2}{c}{\textbf{Fixed $\operatorname{Sim}(3)$}} & \multicolumn{2}{c}{\textbf{Refit $\operatorname{Sim}(3)$}} \\
\cmidrule(lr){4-5} \cmidrule(lr){6-7}
\textbf{Geometry} & \textbf{Benchmark} & \textbf{Current} & GT query geometry & GT motion & GT query geometry & GT motion \\
\midrule
\multirow{4}{*}{VGGT-$\Omega$}
& PStudio         & 0.0947 / 0.9155 & 0.0900 / 0.9169 & 0.0900 / 0.9239 & 0.0763 / 0.9415 & 0.0672 / 0.9492 \\
& PointOdyssey    & 0.4127 / 0.6668 & 0.0531 / 0.9450 & 0.4035 / 0.6787 & 0.0567 / 0.9518 & 0.4017 / 0.6810 \\
& Dynamic Replica & 0.2257 / 0.7777 & 0.0275 / 0.9743 & 0.2207 / 0.7850 & 0.0344 / 0.9764 & 0.2202 / 0.7851 \\
& ADT (secondary) & 0.0933 / 0.9230 & 0.0374 / 0.9582 & 0.1056 / 0.9105 & 0.0261 / 0.9745 & 0.0789 / 0.9459 \\
\midrule
\multirow{4}{*}{DAv3-L}
& PStudio         & 0.1270 / 0.8625 & 0.1146 / 0.8761 & 0.1289 / 0.8614 & 0.0933 / 0.9147 & 0.1011 / 0.9050 \\
& PointOdyssey    & 0.3536 / 0.6616 & 0.0970 / 0.8944 & 0.3418 / 0.6719 & 0.0994 / 0.8979 & 0.3207 / 0.6980 \\
& Dynamic Replica & 0.2836 / 0.7182 & 0.0337 / 0.9666 & 0.2806 / 0.7230 & 0.0426 / 0.9682 & 0.2799 / 0.7252 \\
& ADT (secondary) & 0.1400 / 0.8498 & 0.0555 / 0.9401 & 0.1508 / 0.8388 & 0.0444 / 0.9615 & 0.1151 / 0.8834 \\
\bottomrule
\end{tabular*}
\end{table}

\begin{table}[!htbp]
\centering
\captionsetup{font=footnotesize}
\caption{\textbf{Exact squared-error budget of \nickname-Reg.}
Sequence-macro mean squared error in m$^2$ under the original $\operatorname{Sim}(3)$, decomposed as in \cref{eq:mse-identity}; the identity holds to floating-point precision.
The cross term is large and negative on PStudio, so anchor and motion errors must not be read as additive EPE contributions.}%
\label{tab:mse-budget}
\renewcommand{\arraystretch}{1.05}
\setlength{\tabcolsep}{2pt}
\fontsize{8}{10}\selectfont
\begin{tabular*}{\linewidth}{@{\extracolsep{\fill}}ll cccc@{}}
\toprule
\textbf{Geometry} & \textbf{Benchmark} & \textbf{Total} & \textbf{Anchor} & \textbf{Motion} & \textbf{Cross} \\
\midrule
\multirow{3}{*}{VGGT-$\Omega$}
& PStudio         & 0.0219 & 0.0197 & 0.0200 & $-0.0178$ \\
& PointOdyssey    & 0.6349 & 0.6299 & 0.0127 & $-0.0076$ \\
& Dynamic Replica & 0.2103 & 0.2074 & 0.0060 & $-0.0031$ \\
\midrule
\multirow{3}{*}{DAv3-L}
& PStudio         & 0.0307 & 0.0335 & 0.0280 & $-0.0308$ \\
& PointOdyssey    & 0.3784 & 0.3658 & 0.0422 & $-0.0297$ \\
& Dynamic Replica & 0.2560 & 0.2558 & 0.0090 & $-0.0088$ \\
\bottomrule
\end{tabular*}
\end{table}

\Cref{tab:oracle-full} reports the substitutions discussed in \cref{sec:regression-results} on all primary benchmarks and both geometry backends, together with the secondary ADT benchmark and with refitted scores.
After applying the original $\operatorname{Sim}(3)$ $S$ to the prediction, write $p = S(\hat{P})$, $a = p_0 - G_0$ for the anchor error at the query frame and $m = (p_j - p_0) - (G_j - G_0)$ for the relative-motion error.
\par\begin{samepage}
For every retained observation,
\begin{equation}
\label{eq:mse-identity}
\|p_j - G_j\|^2 = \|a + m\|^2 = \|a\|^2 + \|m\|^2 + 2\,a^{\!\top} m,
\end{equation}
which \cref{tab:mse-budget} reports as sequence-macro means.
\end{samepage}
On PointOdyssey and Dynamic Replica the anchor term dominates the budget for both geometry backends; on PStudio the two terms are comparable and the negative cross term cancels a substantial part of either, so a motion oracle can even raise the fixed-alignment EPE (DAv3-L).
The geometry oracle remains backend-dependent: PointOdyssey EPE is $0.053$ with VGGT-$\Omega$ and $0.097$ with DAv3-L\@.

\paragraph{Swapping the geometry backend.}
The same conclusion follows from replaying the scorecard checkpoint with STream3R~\cite{stream3r2025} point maps computed from $256$-pixel-short-side frames, the scaffold of \cref{tab:track_result}, under the unchanged protocol (\cref{tab:regression}).
PointOdyssey EPE halves ($0.413 \to 0.208$, APD $0.667 \to 0.805$) with no change to the tracker, Dynamic Replica changes little ($0.226 \to 0.228$), and PStudio and ADT get worse ($0.095 \to 0.131$, $0.093 \to 0.117$); the official TrackCraft3R checkpoint shows the same pattern on PointOdyssey ($0.437 \to 0.223$).
No single backend is best on every benchmark, and the pattern is not specific to the reported weights: \cref{tab:backend-steps} repeats the swap for two earlier checkpoints of the lineage with the same outcome.
These are backend-conditioned comparisons of complete systems: they show that performance depends on the supplied geometry, not that the anchor alone explains every difference between backends.
Since the closed-form $\operatorname{Sim}(3)$ minimizes squared distances rather than EPE or APD, a refit can score a higher EPE than the fixed alignment, as on PointOdyssey and Dynamic Replica; neither is a bound, and we report both.

\begin{table}[htbp]
\centering
\captionsetup{font=footnotesize}
\caption{\textbf{Backend swap across checkpoints of the \nickname-Reg lineage.}
EPE$\downarrow$ / APD$\uparrow$ on all points using VGGT-$\Omega$ or STream3R point maps, with fixed weights within each checkpoint. The reported checkpoint follows \cref{tab:regression}; inference settings for the earlier checkpoints have not been fully verified, so cross-checkpoint differences are descriptive.
Across checkpoints, STream3R roughly halves PointOdyssey EPE; Dynamic Replica changes little, while PStudio and ADT prefer VGGT-$\Omega$.
}%
\label{tab:backend-steps}
\renewcommand{\arraystretch}{1.05}
\setlength{\tabcolsep}{2pt}
\fontsize{8}{10}\selectfont
\begin{tabular*}{\linewidth}{@{\extracolsep{\fill}}ll cccc@{}}
\toprule
\textbf{Checkpoint} & \textbf{Geometry} & \textbf{PStudio} & \textbf{PointOdyssey} & \textbf{Dynamic Replica} & \textbf{ADT} \\
\midrule
\multirow{2}{*}{parent run, step $13{,}750$}
& VGGT-$\Omega$ & 0.2317 / 0.7197 & 0.4337 / 0.6426 & 0.2436 / 0.7550 & 0.1030 / 0.9106 \\
& STream3R      & 0.2420 / 0.7082 & 0.2350 / 0.7693 & 0.2476 / 0.7611 & 0.1234 / 0.8764 \\
\midrule
\multirow{2}{*}{recovery stage, step $7{,}500$}
& VGGT-$\Omega$ & 0.0988 / 0.9080 & 0.4146 / 0.6648 & 0.2272 / 0.7758 & 0.0995 / 0.9157 \\
& STream3R      & 0.1338 / 0.8527 & 0.2103 / 0.8016 & 0.2300 / 0.7848 & 0.1183 / 0.8841 \\
\midrule
\multirow{2}{*}{recovery stage, step $1{,}000$ (reported)}
& VGGT-$\Omega$ & 0.0947 / 0.9155 & 0.4127 / 0.6668 & 0.2257 / 0.7777 & 0.0933 / 0.9230 \\
& STream3R      & 0.1311 / 0.8572 & 0.2078 / 0.8050 & 0.2284 / 0.7871 & 0.1168 / 0.8854 \\
\bottomrule
\end{tabular*}
\end{table}

\paragraph{Selected ADT failures of TrackCraft3R.}
The large DAv3-L/ADT gap in \cref{tab:regression} was examined on three sequences, two without dynamic queries and one mixed.
Independently aligned query errors of the two systems are comparable, but the TrackCraft3R replay produces much larger within-window displacement for static queries; removing its per-window query correction does not change its fixed-alignment EPE materially, so the spurious motion lives in the temporal output rather than in the anchor.
This locates the behavior; it does not identify the VAE, the DiT or the decoder as its cause, since no component-level ablation was run.
A residual numerical difference between the historical predictions and an A100 replay (largest selected-case EPE change $0.00035$\,m) remains unresolved and is small relative to the performance gap above.

\subsection{Source and spatial distribution of geometric artifacts}%
\label{appendix:artifact-analysis}

We investigate three possible sources of the flying-point artifacts: the point-map scaffold, \rtv{} residual prediction, and the video VAE codec. All diagnostics use the same fixed, scene-disjoint subset of Kubric val10 and evaluate outputs in metric 3D space without $\operatorname{Sim}(3)$ alignment, so their absolute values are not comparable with \cref{tab:track_result}. For the spatial analysis, edge pixels are the top $10\%$ of ground-truth local 3D discontinuities, dilated by two pixels.

\paragraph{Controlled stage-wise diagnostics.}
We replace the other stages with ground-truth inputs to estimate the error associated with each component (\cref{tab:stage-diagnostics}). These diagnostics do not form an additive decomposition: stages interact, and predicted residuals can partly correct scaffold errors.

\begin{table}[!htbp]
\centering
\captionsetup{font=footnotesize}
\caption{\textbf{Controlled stage-wise diagnostics of geometric error.}
Metric-space EPE on a fixed Kubric val10 subset. The arms isolate different sources and are not additive.}
\label{tab:stage-diagnostics}
\renewcommand{\arraystretch}{1.05}
\setlength{\tabcolsep}{2pt}
\fontsize{8}{10}\selectfont
\begin{tabular*}{\linewidth}{@{\extracolsep{\fill}}lcc@{}}
\toprule
\textbf{Diagnostic arm} & \textbf{Isolates} & \textbf{All EPE $\downarrow$} \\
\midrule
Predicted point-map scaffold & scaffold error & $\approx2.34$ \\
Oracle scaffold + predicted RTV residual & residual-prediction error & $\approx0.84$ \\
Ground-truth target $\rightarrow$ VAE round-trip & VAE codec floor & $\approx0.79$ \\
Full pipeline & all stages combined & $\approx2.12$ \\
\bottomrule
\end{tabular*}
\end{table}

The scaffold has the largest isolated error. With an oracle scaffold, the predicted residual contributes approximately $0.84$ EPE, while a VAE round-trip of the ground-truth target produces an approximately $0.79$ EPE floor. The full pipeline slightly improves over the predicted scaffold alone, consistent with the residual correcting some scaffold error.

\paragraph{Boundary localization and metric impact.}
Passing ground-truth targets through the VAE without diffusion yields approximately $2.5\times$ higher reconstruction error near ground-truth 3D discontinuities than in interior regions. We further compare the VAE-decoded model with a data-matched direct-readout model trained on the same Kubric tracking data, without auxiliary depth/camera supervision or PointOdyssey (\cref{tab:boundary-diagnostics}). Both rows of \cref{tab:boundary-diagnostics} are these data-matched training runs, so their absolute EPE differs from the full-pipeline arm of \cref{tab:stage-diagnostics}.

\begin{table}[!htbp]
\centering
\captionsetup{font=footnotesize}
\caption{\textbf{Metric impact of removing the VAE output path.}
Metric-space EPE on the fixed Kubric val10 subset.}
\label{tab:boundary-diagnostics}
\renewcommand{\arraystretch}{1.05}
\setlength{\tabcolsep}{2pt}
\fontsize{8}{10}\selectfont
\begin{tabular*}{\linewidth}{@{\extracolsep{\fill}}lccc@{}}
\toprule
\textbf{Output path} & \textbf{All EPE $\downarrow$} & \textbf{Edge EPE $\downarrow$} & \textbf{Interior EPE $\downarrow$} \\
\midrule
VAE-decoded & 1.340 & 1.479 & 1.300 \\
VAE-free direct readout & \textbf{0.859} & \textbf{1.331} & \textbf{0.728} \\
Relative reduction & 35.9\% & 10.0\% & 44.0\% \\
\bottomrule
\end{tabular*}
\end{table}

In this data-matched comparison, replacing the VAE-decoded output path with direct readout reduces overall EPE by $35.9\%$; with the round-trip floor above, this indicates that the codec meaningfully affects the reported metric. However, the gain is much larger in interior regions ($44.0\%$) than at boundaries ($10.0\%$), and the direct-readout model still has approximately $1.8\times$ higher edge EPE than interior EPE. The evidence therefore identifies the scaffold as the main source of global geometric error and the VAE as a secondary error floor that accentuates boundary-localized artifacts; neither the VAE nor residual prediction alone explains all flying points.

\subsection{Camera-following-object failure analysis}%
\label{appendix:camera-following-analysis}

The failure case in \cref{fig:supp-failurecase} raises the possibility that the log-space transform in \cref{eq:track_offset_normalization}, including the scale factor $K$, compresses genuine deformation toward zero. We audit this hypothesis on $50$ Kubric clips in which the camera follows a moving object, comparing the true world-space displacement with the normalized target. Substantial motion remains clearly separated from zero after normalization; for example, a clip with approximately $9$\,m of object displacement retains a strongly non-zero normalized target. The audit therefore does not support the specific hypothesis that log-space normalization universally erases the deformation signal.

This result does not by itself prove dataset bias. Because the generative model receives dynamic supervision only from our Kubric subset, while other 3D trackers combine Kubric with datasets such as Dynamic Replica and PointOdyssey, insufficient coverage of camera-following-object trajectories is a plausible explanation. This hypothesis remains unverified. Identifying the precise stage at which the model suppresses the preserved motion signal requires tracing the normalized target, model prediction, and decoded output on the same clips.

\subsection{Positioning relative to recent feed-forward methods}%
\label{appendix:recent-methods}

Several recent methods approach dense dynamic reconstruction through direct feed-forward geometric regression. POMATO~\cite{zhang25pomato:} combines point-map matching with temporal motion in a DUSt3R-style framework. Zero-MSF~\cite{liang25zeromsf} jointly predicts geometry and scene flow and scales training to approximately one million annotated synthetic samples. Track4World~\cite{lu26track4world} uses a VGGT-style video encoder and 3D correlation to regress pairwise dense 2D/3D flows that are composed into world-centric tracks. The concurrent D4RT~\cite{zhang2025d4rt} encodes a video into a global representation and decodes flexible spatio-temporal point queries, while the concurrent 4RC~\cite{luo264rc:} encodes a video once and conditionally queries dense geometry and motion at arbitrary times.

These deterministic regressors are complementary to \nickname{}. They optimize geometric outputs directly, whereas our primary question is how a pretrained generative video prior can represent and produce dense 3D tracking signals. The direct-readout experiment in \cref{appendix:direct-readout} begins to connect the two categories by retaining the generative video backbone while replacing the RGB-oriented output codec with a geometric regression head. Concurrent work on general-purpose perception from video-generative pretraining~\cite{wang26genception} and omnimodal world models that combine understanding, generation, simulation, and action~\cite{nvidia26cosmos3} further motivates studying tracking within shared generative backbones. A similar distinction exists in depth estimation, where generative approaches~\cite{ke2024repurposing,xu2025pixel} are studied as a complementary category alongside mature discriminative regressors. The performance gap to mature specialist regressors remains an important limitation.

\section{More results and discussion}%
\label{appendix:more}

\begin{figure}[h!]
\centering
\includegraphics[width=\textwidth]{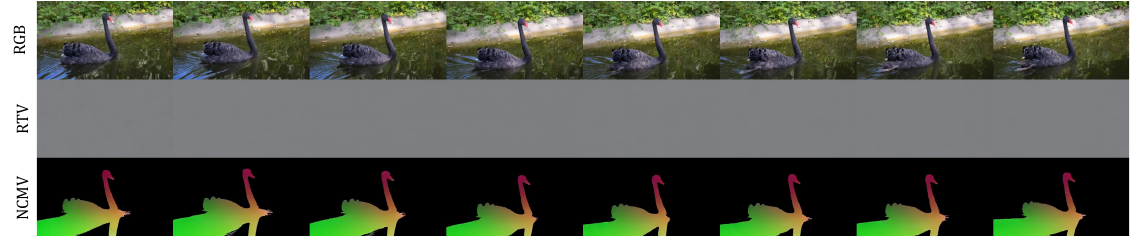}
\caption{\textbf{Failure case.}
\nickname{} predicts near-zero deformation in this video, where the camera moves with the object.
Limited coverage of such camera-following-object trajectories in the training data is a plausible, but not conclusively established, cause (\cref{appendix:camera-following-analysis}).
}%
\label{fig:supp-failurecase}
\end{figure}
\renewcommand{\topfraction}{0.95}\renewcommand{\bottomfraction}{0.95}\renewcommand{\textfraction}{0.001}\renewcommand{\floatpagefraction}{0.8}\setcounter{totalnumber}{4}\setcounter{topnumber}{3}

\paragraph{Decoding tracks from \ncmv: CoTracker vs. nearest-neighbor.}%
\label{appendix:nn}

Replacing CoTracker with nearest-neighbor color matching and a Kalman filter reduces APD from $0.4798$ to $0.3174$ on PointOdyssey and from $0.7495$ to $0.5052$ on PStudio. Thus, finite-resolution \ncmv{} colors alone are insufficient for reliable matching; the 2D tracker materially improves decoding.

\begin{figure}[p]
\centering
\includegraphics[width=\textwidth]{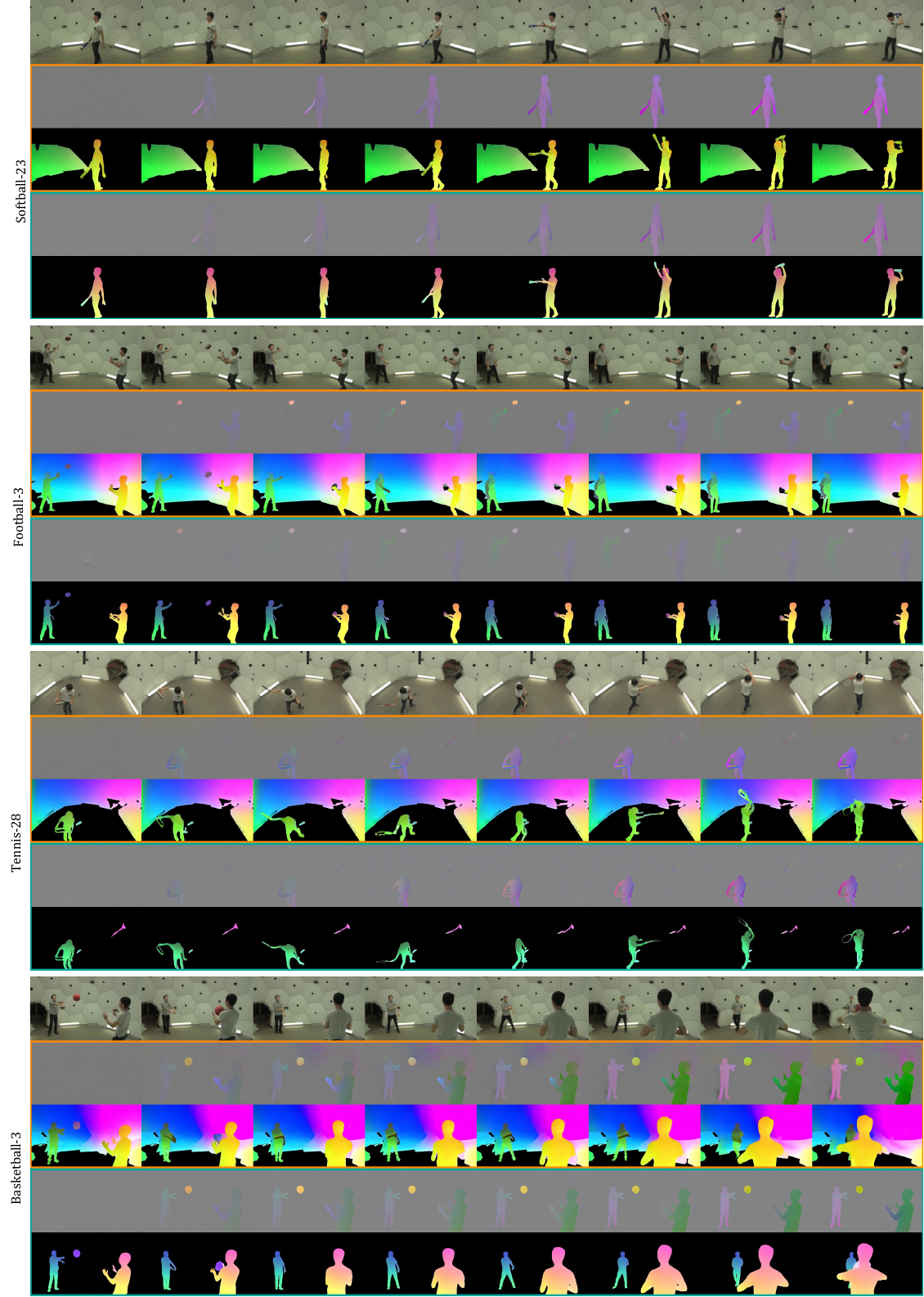}
\caption{\textbf{Effect of backbone size.}
Specifically, we show our model's performance on four samples from PStudio~\cite{pstudio}, with sequence names on the left.
For each sequence, we compare the results from \nickname-5B (in \corange{orange}) and \nickname-14B (in \cteal{green}).
Both \rtv{} and \ncmv{} are visualized.
The Wan-14B-based \nickname{} shows better foreground-background factorization and sharper motion, suggesting benefits from stronger video diffusion models.}%
\label{fig:supp-5b_vs_14b}
\end{figure}

\begin{figure}[p]
\centering
\includegraphics[width=\textwidth]{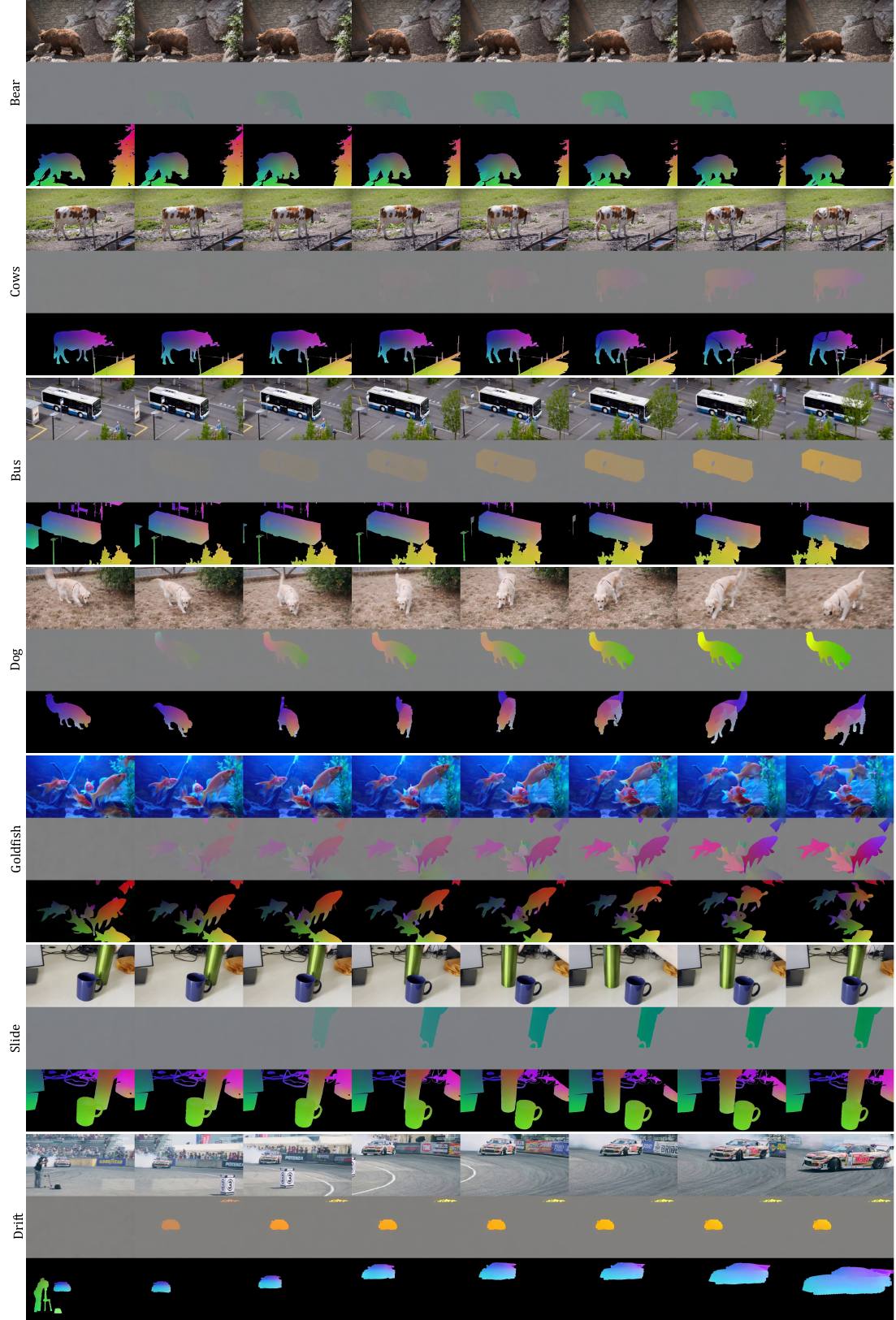}
\caption{\textbf{More qualitative results.}
We show seven zero-shot inference results from DAVIS~\cite{davis_CVPR_2016}.
The first row shows RGB input, with eight frames sampled uniformly from a $49$-frame clip, and the second and third rows show \rtv{} and \ncmv{} predictions.
The query frame is the first frame shown in the first row.
\rtv{} encodes 3D offsets on the query-frame grid; \ncmv{} encodes persistent coordinates on each video frame's grid.}%
\label{fig:supp-more-results}
\end{figure}

\paragraph{Additional visualizations.}
\Cref{fig:supp-more-results} shows additional zero-shot DAVIS results; \cref{fig:supp-5b_vs_14b} compares Wan 5B and 14B on the same examples, where 14B produces sharper motion.

\begin{figure}[t]
\centering
\includegraphics[width=1.0\textwidth]{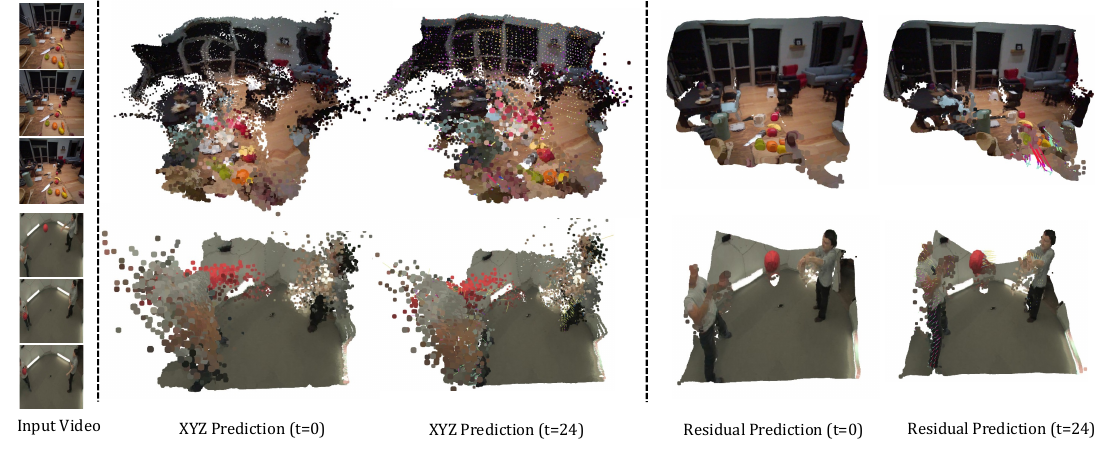}
\caption{\textbf{Ablation: absolute XYZ prediction vs.\ residual prediction.}
We visualize 3D tracking results at $t=0$ and $t=24$ on ADT~\cite{pan2023ariadigitaltwinnew} and PStudio~\cite{pstudio} sequences.
Left: predicting absolute 3D coordinates directly (\textit{w/o Residual Pred}).
Right: our residual formulation (\cref{eq:track_offset}).
Absolute prediction produces noisy geometry with scattered outliers, whereas the residual formulation yields cleaner reconstructions with more coherent motion.
}%
\label{fig:ablation_residual}
\end{figure}


\FloatBarrier
\section{Broader impacts}
\label{appendix:impacts}
Dense 3D tracking from monocular video can support robotics, 4D reconstruction, video editing, and scientific motion analysis.
Like other motion-tracking methods, it could also be used to follow people in recorded video; \nickname is trained on synthetic and public research datasets, but improved tracking can still increase privacy risks.
Its outputs are estimates conditioned on an external reconstruction and can be wrong, particularly for content that is occluded, enters after the query frame, or moves with the camera (\cref{appendix:camera-following-analysis}); they require validation before use as measurements.

\section{Existing assets and licenses}
\label{appendix:licenses}
\Cref{tab:asset_licenses} lists the external datasets, models and baselines, how we use them, and their licenses. Code, the adapters and the re-rendered Kubric split will be released at \mbox{\url{https://github.com/NIRVANALAN/Track4D}}.
\begin{table}[!htbp]
\centering
\captionsetup{font=footnotesize}
\caption{\textbf{Existing assets and their licenses}, as stated by the official repositories and dataset pages (checked October 2026). Data and model weights may carry terms beyond the code license; we use every asset for non-commercial research only.}
\label{tab:asset_licenses}
\fontsize{8}{10}\selectfont
\renewcommand{\arraystretch}{1.05}
\setlength{\tabcolsep}{3pt}
\begin{tabular*}{\linewidth}{@{\extracolsep{\fill}}>{\raggedright\arraybackslash}p{0.2\linewidth}>{\raggedright\arraybackslash}p{0.3\linewidth}>{\raggedright\arraybackslash}p{0.42\linewidth}@{}}
\toprule
\textbf{Asset} & \textbf{Use} & \textbf{License / terms} \\
\midrule
Kubric~\cite{greff2021kubric} & Rendering our training split & Apache-2.0 (code) \\
TartanAir~\cite{tartanair2020iros} & Training (generative) & CC BY 4.0 \\
PointOdyssey~\cite{zheng2023point} & Training (regression), evaluation & CC BY-NC-SA 4.0 (data); MIT (code) \\
Dynamic Replica~\cite{karaev2023dynamicstereo} & Training (regression), evaluation & CC BY-NC (repository); data after accepting the license agreement \\
Waymo Open Dataset~\cite{waymo} & Training (regression) & Waymo Open Dataset terms; Apache-2.0 (code) \\
DL3DV-10K~\cite{ling2024dl3dv} & Training (regression) & DL3DV-10K terms of use and CC BY-NC 4.0 \\
ScanNet~\cite{dai2017scannet} & Training (regression) & ScanNet Terms of Use; MIT (code) \\
MVS-Synth~\cite{DeepMVS} & Training (regression) & Official terms; BSD-2-Clause (DeepMVS code) \\
ADT, PStudio (TAPVid-3D) & Evaluation & Aria Digital Twin license and Panoptic Studio terms; Apache-2.0 (TAPVid-3D code) \\
DAVIS~\cite{davis_CVPR_2016} & Qualitative results & Official terms \\
\midrule
Wan2.1, Wan2.2~\cite{wan2025} & Backbones & Apache-2.0 \\
VGGT-$\Omega$~\cite{wang26vggt-omega} & Input geometry & FAIR Noncommercial Research License \\
STream3R~\cite{stream3r2025} & Input geometry & S-Lab License 1.0 (non-commercial) \\
Depth Anything 3~\cite{lin2026depth} & Input geometry & Apache-2.0 (code); CC BY-NC 4.0 (Large/Giant weights) \\
CoTracker~\cite{cotracker3} & \ncmv{} decoding, baseline & CC BY-NC 4.0 \\
TrackCraft3R~\cite{nam2026trackcraft3r} & Baseline & Apache-2.0 \\
V-DPM~\cite{sucar2025vdpm} & Baseline & MIT \\
4RC~\cite{luo264rc:} & Baseline & S-Lab License 1.0 (non-commercial) \\
Any4D~\cite{karhade2025any4d} & Baseline & Apache-2.0 \\
TraceAnything~\cite{liu2025traceanythingrepresentingvideo} & Baseline & Apache-2.0 \\
SpaTrackerV2~\cite{xiao2025spatialtracker} & Baseline & CC BY-NC 4.0 \\
St4RTrack~\cite{st4rtrack2025} & Baseline & No license file in the official repository \\
\bottomrule
\end{tabular*}
\end{table}

\clearpage
\section*{NeurIPS Paper Checklist}

\begin{enumerate}

\Needspace{7\baselineskip}

\item {\bf Claims}
    \item[] Question: Do the main claims made in the abstract and introduction accurately reflect the paper's contributions and scope?
    \item[] Answer: \answerYes{}
    \item[] Justification: The abstract and \cref{sec:intro} state the contributions: a study of how to represent dense 3D tracking inside a pretrained video diffusion model, the comparison of the \rtv{} and \ncmv{} readouts, and a regression readout of the same residual without the video VAE. They are supported by \cref{tab:track_result,tab:regression,tab:ablation_progressive,tab:rtv-only-ablation} and \cref{sec:regression-results}; the scope (49-frame clips, point maps from an off-the-shelf reconstructor, frame-0 queries) is stated in \cref{sec:method,sec:experiments}.

\Needspace{7\baselineskip}

\item {\bf Limitations}
    \item[] Question: Does the paper discuss the limitations of the work performed by the authors?
    \item[] Answer: \answerYes{}
    \item[] Justification: \Cref{sec:conclusions} and \cref{appendix:limitations} state the limitations: the dense-target requirement of the generative readouts, the fixed 49-frame window, frame-0 queries, and the dependence on the supplied geometry (quantified by the oracle and backend swaps of \cref{appendix:oracle}). The appendix analyses the VAE boundary artifacts (\cref{appendix:artifact-analysis}), the camera-following failure (\cref{appendix:camera-following-analysis}, \cref{fig:supp-failurecase}) and the inference cost (\cref{appendix:efficiency-comparison}).

\Needspace{7\baselineskip}

\item {\bf Theory assumptions and proofs}
    \item[] Question: For each theoretical result, does the paper provide the full set of assumptions and a complete (and correct) proof?
    \item[] Answer: \answerNA{}
    \item[] Justification: The paper is empirical and presents no theorems requiring formal proof.

    \Needspace{7\baselineskip}

    \item {\bf Experimental result reproducibility}
    \item[] Question: Does the paper fully disclose all the information needed to reproduce the main experimental results of the paper to the extent that it affects the main claims and/or conclusions of the paper (regardless of whether the code and data are provided or not)?
    \item[] Answer: \answerYes{}
    \item[] Justification: \Cref{sec:method} defines the representations and \cref{sec:regression} the regression readout. \Cref{appendix:data,appendix:arch} give the training data, LoRA configuration, optimizer, batch size, iterations and hardware of the generative model; \cref{appendix:regression} gives the architecture, normalization, training mixture, schedule and the 49-frame protocol of \nickname-Reg; \cref{appendix:protocol} gives the comparison protocol of every baseline.

\Needspace{7\baselineskip}

\item {\bf Open access to data and code}
    \item[] Question: Does the paper provide open access to the data and code, with sufficient instructions to faithfully reproduce the main experimental results, as described in supplemental material?
    \item[] Answer: \answerYes{}
    \item[] Justification: The LoRA adapters are public at \url{https://huggingface.co/yslan/Track4D-2026}; the code and the re-rendered Kubric split will be released at \url{https://github.com/NIRVANALAN/Track4D}. The appendix describes data, training and evaluation in enough detail for re-implementation.

\Needspace{7\baselineskip}

\item {\bf Experimental setting/details}
    \item[] Question: Does the paper specify all the training and test details (e.g., data splits, hyperparameters, how they were chosen, type of optimizer) necessary to understand the results?
    \item[] Answer: \answerYes{}
    \item[] Justification: \Cref{appendix:arch} specifies LoRA rank, learning rate, batch size, training length, inference noise schedule, adapted layers, frozen modules and domain tokens of the generative model; \cref{appendix:regression} gives the same for \nickname-Reg, including the training mixture and the three-stage warm-start lineage.
    \Cref{appendix:protocol} documents the protocol used for each baseline.

\Needspace{7\baselineskip}

\item {\bf Experiment statistical significance}
    \item[] Question: Does the paper report error bars suitably and correctly defined or other appropriate information about the statistical significance of the experiments?
    \item[] Answer: \answerNo{}
    \item[] Justification: Following common practice in 3D point tracking benchmarks, we report deterministic APD/EPE averaged over $50$ sequences per benchmark ($49$ for ADT in \cref{tab:regression}) without error bars. The joint-versus-\rtv-only comparison uses one checkpoint pair with different training histories and sampling settings, and claims no significance (\cref{appendix:ncmv-analysis}).

\Needspace{7\baselineskip}

\item {\bf Experiments compute resources}
    \item[] Question: For each experiment, does the paper provide sufficient information on the computer resources (type of compute workers, memory, time of execution) needed to reproduce the experiments?
    \item[] Answer: \answerYes{}
    \item[] Justification: The joint Wan2.1-I2V-14B model underwent $5{,}000$ joint-training iterations on $32$ NVIDIA GH200 GPUs for about one day (about $770$ GPU-hours; \cref{appendix:arch}); \nickname-Reg was trained on two NVIDIA H200 GPUs for about seven days over its three stages (about $340$ GPU-hours; \cref{appendix:regression}). Inference cost is reported in \cref{appendix:efficiency-comparison,appendix:sampling_steps}: about $14$\,s per $49$-frame clip for the sampling-ablation model at $50$ steps on an A6000, and $6.5$\,s with $26.0$\,GB reserved memory for \nickname-Reg on an A100, excluding the input point maps (\cref{tab:runtime-comparison}).

\Needspace{7\baselineskip}

\item {\bf Code of ethics}
    \item[] Question: Does the research conducted in the paper conform, in every respect, with the NeurIPS Code of Ethics \url{https://neurips.cc/public/EthicsGuidelines}?
    \item[] Answer: \answerYes{}
    \item[] Justification: We have reviewed the NeurIPS Code of Ethics. The paper uses publicly released datasets and pretrained models, involves no human-subject studies, and evaluates tracking on existing datasets.

\Needspace{7\baselineskip}

\item {\bf Broader impacts}
    \item[] Question: Does the paper discuss both potential positive societal impacts and negative societal impacts of the work performed?
    \item[] Answer: \answerYes{}
    \item[] Justification: \Cref{appendix:impacts} discusses positive and negative impacts. The paper proposes a research method for 3D motion estimation, with applications in robotics, video understanding and 4D reconstruction. As with any motion-tracking technology, downstream uses could include surveillance; improved tracking can increase privacy risks even without a specific deployment.

\Needspace{7\baselineskip}

\item {\bf Safeguards}
    \item[] Question: Does the paper describe safeguards that have been put in place for responsible release of data or models that have a high risk for misuse (e.g., pre-trained language models, image generators, or scraped datasets)?
    \item[] Answer: \answerNA{}
    \item[] Justification: The released artifacts are a LoRA adapter for a publicly available video diffusion model and a re-rendering of a synthetic dataset (Kubric). They do not pose a high misuse risk beyond that already present in the underlying public assets.

\Needspace{7\baselineskip}

\item {\bf Licenses for existing assets}
    \item[] Question: Are the creators or original owners of assets (e.g., code, data, models), used in the paper, properly credited and are the license and terms of use explicitly mentioned and properly respected?
    \item[] Answer: \answerYes{}
    \item[] Justification: \Cref{tab:asset_licenses} (\cref{appendix:licenses}) lists every external asset with its license. All datasets and pretrained models (Kubric, TartanAir, PointOdyssey, Dynamic Replica, Waymo, DL3DV, ScanNet, MVS-Synth, Aria Digital Twin, Panoptic Studio, DAVIS, Wan2.1, Wan2.2, VGGT-$\Omega$, STream3R, Depth Anything~3, CoTracker, and the baselines' official checkpoints) are cited at first use and used under their released research licenses.

\Needspace{7\baselineskip}

\item {\bf New assets}
    \item[] Question: Are new assets introduced in the paper well documented and is the documentation provided alongside the assets?
    \item[] Answer: \answerYes{}
    \item[] Justification: We will release at \url{https://github.com/NIRVANALAN/Track4D} the re-rendered Kubric split with dense 3D tracking annotations, the rendering scripts, and the trained adapters of the generative and regression models, with documentation of the data format, training commands and inference protocol of \cref{appendix:arch,appendix:regression}.

\Needspace{7\baselineskip}

\item {\bf Crowdsourcing and research with human subjects}
    \item[] Question: For crowdsourcing experiments and research with human subjects, does the paper include the full text of instructions given to participants and screenshots, if applicable, as well as details about compensation (if any)?
    \item[] Answer: \answerNA{}
    \item[] Justification: The work involves no crowdsourcing or research with human subjects.

\Needspace{7\baselineskip}

\item {\bf Institutional review board (IRB) approvals or equivalent for research with human subjects}
    \item[] Question: Does the paper describe potential risks incurred by study participants, whether such risks were disclosed to the subjects, and whether Institutional Review Board (IRB) approvals (or an equivalent approval/review based on the requirements of your country or institution) were obtained?
    \item[] Answer: \answerNA{}
    \item[] Justification: The work involves no human subjects, so IRB review is not applicable.

\Needspace{7\baselineskip}

\item {\bf Declaration of LLM usage}
    \item[] Question: Does the paper describe the usage of LLMs if it is an important, original, or non-standard component of the core methods in this research? Note that if the LLM is used only for writing, editing, or formatting purposes and does \emph{not} impact the core methodology, scientific rigor, or originality of the research, declaration is not required.
    \item[] Answer: \answerNA{}
    \item[] Justification: LLMs are not part of the core method. They were used only as writing assistants for editing and copy-editing, which does not require declaration under the NeurIPS LLM policy.

\end{enumerate}
\end{document}